\documentclass[10pt,twocolumn,letterpaper]{article}

\usepackage[pagenumbers]{cvpr} 

\usepackage{amsmath}
\usepackage{threeparttable}
\usepackage{adjustbox}
\usepackage{amssymb}
\usepackage[utf8]{inputenc} 
\usepackage[T1]{fontenc}    
\usepackage{url}            
\usepackage{booktabs}       
\usepackage{amsfonts}       
\usepackage{nicefrac}       
\usepackage{microtype}      
\usepackage{xcolor}         
\usepackage{algorithm}

\usepackage[numbers,sort&compress]{natbib}
\usepackage{graphicx}
\usepackage{dcolumn}
\usepackage{subcaption}
\usepackage{multirow}
\usepackage{multicol}
\usepackage{enumitem}
\usepackage{bm}
\usepackage{algpseudocode}
\usepackage{wrapfig,lipsum,booktabs}

\usepackage{float}
\usepackage{subcaption}
\usepackage{tabularx}

\usepackage{pgfplots}
\usepackage{graphicx}
\usepackage{xcolor}
\usepackage{pgf-pie}  
\pgfplotsset{compat=1.16}

\usepackage{placeins}

\DeclareMathOperator*{\argmax}{arg\,max}
\newcommand{\norm}[1]{\left\lVert #1 \right\rVert}
\newcommand{\Ls}{\mathcal{L}}
\newcommand{\vx}{\bm{x}}
\newcommand{\vp}{\bm{p}}
\newcommand{\vz}{\bm{z}}
\newcommand{\sign}{\operatorname{sign}}

\definecolor{cvprblue}{rgb}{0.21,0.49,0.74}
\usepackage[pagebackref,breaklinks,colorlinks,allcolors=cvprblue]{hyperref}

\def\paperID{12989} 
\def\confName{CVPR}
\def\confYear{2025}

\title{Towards Million-Scale Adversarial Robustness Evaluation With Stronger Individual Attacks}

\author{
Yong Xie\textsuperscript{1},  
Weijie Zheng\textsuperscript{1},  
Hanxun Huang\textsuperscript{2},  
Guangnan Ye\textsuperscript{1}\textsuperscript{$*$},  
Xingjun Ma\textsuperscript{1}\thanks{Corresponding authors: {\tt\small yegn@fudan.edu.cn, xingjunma@fudan.edu.cn}} \\
\textsuperscript{1}Shanghai Key Lab of Intell. Info. Processing, School of CS, Fudan University, China \\
\textsuperscript{2}School of Computing and Information Systems, The University of Melbourne, Australia
}

\begin{document}
\maketitle

\begin{abstract}
As deep learning models are increasingly deployed in safety-critical applications, evaluating their vulnerabilities to adversarial perturbations is essential for ensuring their reliability and trustworthiness. Over the past decade, a large number of white-box adversarial robustness evaluation methods (i.e., attacks) have been proposed, ranging from single-step to multi-step methods and from individual to ensemble methods. 
Despite these advances, challenges remain in conducting meaningful and comprehensive robustness evaluations, particularly when it comes to large-scale testing and ensuring evaluations reflect real-world adversarial risks.
In this work, we focus on image classification models and propose a novel individual attack method, \emph{Probability Margin Attack} (PMA), which defines the adversarial margin in the probability space rather than the logits space. We analyze the relationship between PMA and existing cross-entropy or logits-margin-based attacks, and show that PMA can outperform the current state-of-the-art individual methods.
Building on PMA, we propose two types of ensemble attacks that balance effectiveness and efficiency. Furthermore, we create a million-scale dataset, CC1M, derived from the existing CC3M dataset, and use it to conduct the first million-scale white-box adversarial robustness evaluation of adversarially-trained ImageNet models. Our findings provide valuable insights into the robustness gaps between individual versus ensemble attacks and small-scale versus million-scale evaluations.
\end{abstract}    
\section{Introduction}
\label{sec:intro}

Despite their success in various applications, deep neural networks (DNNs) exhibit significant sensitivity to imperceptible perturbations specifically designed to maximize prediction error, a phenomenon known as "adversarial vulnerability" \cite{szegedy2013intriguing}. This vulnerability presents a critical safety risk, as adversaries can exploit it to launch stealthy attacks that evade human detection, potentially undermining the reliability of real-world systems. Consequently, evaluating the adversarial robustness of DNNs has become essential.

White-box adversarial attack methods evaluate the worst-case performance of a DNN model by crafting attacks directly using adversarial gradients.
These attacks, commonly tested in image classification tasks, aim to maximize the model's prediction error while constraining perturbations to be "small" through an $\Ls_p$ norm~\cite{goodfellow2014explaining,carlini2017towards}. 
The Projected Gradient Descent (PGD) \cite{madry2017towards} and AutoAttack (AA) \cite{croce2020reliable} are two commonly used for such evaluations. PGD is a strong first-order attack, while AA is an ensemble attack with four different methods. However, PGD struggles with obfuscated gradients \cite{athalye2018obfuscated}, and AA is computationally expensive \cite{ma2023imbalanced}, highlighting the trade-off between effectiveness and efficiency in white-box robustness evaluation.

As large-scale DNNs are increasingly deployed across various applications, large-scale adversarial robustness evaluations have become essential. To conduct such evaluations, highly efficient attack methods are necessary. Although AA can be accelerated through adaptive initialization and discarding strategies \cite{liu2022practical}, individual attacks hold a significant efficiency advantage in large-scale evaluation.
 Improving the standard PGD attack is a promising direction for obtaining stronger individual attacks. Existing improvements to PGD include better initialization \cite{tashiro2020diversity}, multiple adversarial targets \cite{gowal2019alternative}, intermediate feature layers \cite{yu2021lafeat}, or loss alternation \cite{antoniou2022alternating}, as well as more robust loss functions, such as margin loss \cite{ma2023imbalanced}, the difference of logits ratio (DLR) \cite{croce2020reliable}, and Minimize the Impact of Floating-Point Errors (MIFPE) loss \cite{yu2023efficient}. A recent work proposed a Margin Decomposition (MD) attack to circumvent obfuscated/imbalanced gradients using a two-stage margin decomposition strategy.

In this work, we build on the attack pipeline of the Margin Decomposition (MD) attack but introduce a novel loss function, \emph{probability margin loss}, to develop a stronger individual attack. Specifically, the probability margin loss defeines the margin in the probability space, rather than the logits space. We refer to the attack using this adversarial loss function as \textbf{Probability Margin Attack} (PMA). 
Additionally, we explore cost-effective ensemble attacks by combining PMA with other existing attack methods. We also construct a million-scale evaluation dataset, CC1M, derived from the Conceptual Captions 3 Million (CC3M) \cite{sharma2018conceptual}, and use it to conduct a million-scale white-box adversarial robustness evaluation of adversarially trained models on ImageNet.

In summary, our main contributions are:

\begin{itemize}
    \item We propose a novel individual attack \emph{Probability Margin Attack (PMA)}, which introduces a probability margin loss to boost attack effectiveness of individual attacks. We aslo analyze the relationship between probability margin loss and the commonly used cross-entropy and margin losses.
    
    \item We empirically demonstrate that PMA consistently outperforms existing individual attacks across multiple datasets (CIFAR-10, CIFAR-100, and ImageNet-1k) when used to evaluate the top-ranked models on the RobustBench leaderboard \cite{croce2020reliable}. Additionally, we propose two PMA-based ensemble attacks to balance effectiveness with efficiency.
    
    \item We construct a million-scale evaluation dataset, CC1M, consisting of 1 million images derived from the CC3M dataset after removing outliers. Using CC1M, we conduct a large-scale robustness evaluation for adversarially trained ImageNet models, revealing a significant robustness gap between small-scale evaluation on the ImageNet-1k test set and large-scale evaluation on CC1M.
\end{itemize}

\section{Related Work}
\label{sec:related}


\paragraph{White-box Adversarial Attacks}
A number of white-box adversarial attacks have been proposed to assess the robustness of deep neural networks (DNNs). Early white-box attacks include L-BFGS \cite{szegedy2013intriguing}, Fast Gradient Sign Method (FGSM) \cite{goodfellow2014explaining}, and Basic Iterative Method (BIM) \cite{kurakin2016adversarial}. These attacks utilize second-order optimization or single- and multi-step gradient sign updates to generate $L_{\infty}$-bounded adversarial examples. 
Subsequently, optimization-based attacks, such as the Carlini-Wagner (C\&W) attack \cite{carlini2017towards}, were introduced to jointly optimize both the misclassification objective and the perturbation constraint. However, both L-BFGS and C\&W attacks are computationally expensive, while FGSM and BIM, being more efficient, are less effective in generating strong adversarial examples.
To strike a balance between attack strength and computational efficiency, the Projected Gradient Descent (PGD) attack \cite{madry2017towards} was introduced within the adversarial training framework \cite{goodfellow2014explaining,madry2017towards,zhang2019theoretically}. PGD allows the perturbation to exceed the $\epsilon$-ball and uses a clipping operation to ensure that the perturbation remains within the constraint when necessary. Additionally, PGD employs a random initialization strategy to help escape local minima, thereby enhancing its attack strength.


\paragraph{Reliable Adversarial Robustness Evaluation}  

The accuracy of robustness evaluations relies heavily on the strength of the attack. White-box attacks used for robustness evaluation must be resilient to issues such as obfuscated gradients \cite{athalye2018obfuscated} and imbalanced gradients \cite{ma2023imbalanced}. As shown in \cite{athalye2018obfuscated}, standard PGD can overestimate robustness in the presence of obfuscated gradients, leading to the development of methods aimed at improving PGD’s effectiveness. These improvements can be broadly classified into two categories: 1) attack strategies and 2) loss functions.


Improved attack strategies include enhanced initialization \cite{tashiro2020diversity}, adaptive boundaries \cite{croce2020minimally}, multiple target optimization \cite{gowal2019alternative}, and random restarts with step size scheduling \cite{croce2020reliable}. Further strategies such as utilizing intermediate logits \cite{yu2021lafeat}, alternating objectives \cite{antoniou2022alternating}, and margin decomposition \cite{ma2023imbalanced} also contribute to increased robustness. In terms of loss functions, the field has progressed from using cross-entropy to margin loss \cite{carlini2017towards,ma2023imbalanced}, boundary distance loss \cite{croce2020minimally}, DLR \cite{croce2020reliable}, and MIFPE loss \cite{yu2023efficient}. Among these, margin loss has shown notable resilience to gradient-related issues, as demonstrated by the Margin Decomposition (MD) attack \cite{ma2023imbalanced}. The AutoAttack (AA) framework \cite{croce2020reliable} integrates four advanced attack methods—two Auto-PGD (APGD) attacks, a FAB attack \cite{gowal2019alternative}, and a black-box Square attack \cite{andriushchenko2020square}—to provide a strong, ensemble-based approach for robust evaluation.


However, the extensive runtime of AA often exceeds model training times, making robustness evaluation impractical in many cases. This has led to the development of new attack methods aimed at faster and more efficient evaluations. For instance, the LAFEAT attack \cite{yu2021lafeat} enhances PGD by leveraging intermediate feature layers, while Adaptive AA ($A^3$) \cite{liu2022practical} accelerates AA with an adaptive direction initialization strategy. Additionally, the Auto Conjugate Gradient (ACG) attack \cite{yamamura2022diversified} applies the Conjugate Gradient (CG) method to improve adversarial optimization. Recently, the Margin Decomposition (MD) attack \cite{ma2023imbalanced} introduced a two-phase approach that integrates margin loss to address gradient issues and strengthen attacks. In this work, we build upon the MD attack by proposing a simpler but more effective loss function that surpasses existing individual attacks.
\section{Proposed Attack}
\label{sec:PMA}

\paragraph{Preliminary}
In this work, we focus on adversarial attacks on image classification models. Given a target model $f$, the goal of an adversarial attack is to find an adversarial perturbation $\bm{\delta}^*$ constrained within a small $\epsilon$-ball to maximize the model's classification error. Formally, it is:
\begin{equation}
    \bm{\delta}^* = \argmax_{\norm{\bm{\delta}}_{\infty} \leq \epsilon} \Ls(f(\vx+\bm{\delta}), y),
\end{equation}
where $\Ls$ denotes the classification loss, $y$ is the correct label of $\vx$, and $\norm{\cdot}_{\infty}$ is the $L_{\infty}$ norm. The generated adversarial example is denoted by $\vx_{adv} = \vx+\bm{\delta}^*$.

\subsection{Probability Margin Loss}
Let $\vz$ be the logit output of $f(\vx)$ and $\vz_i$ be the logits output of the $i$-th class, and $\vp_i=e^{\overset{\vz_i}{}}/\sum_{i=1}^{N}e^{\overset{\vz_i}{}}$ be the probability output of the $i$-th class for a total number of $N$ classes. Sorting the values of $\vz_i$ in ascending order, $\vz_{\pi_i}$ represents the $i$-th largest logit value (except $\vz_{y}$).
Our proposed \emph{probability margin loss} is defined as: 
\begin{equation}
\Ls_{pm}\left(\vz, y\right) = \vp_{max} - \vp_y = \frac{e^{\vz_{max}}-e^{\vz_y}}{\sum_{i}^{N}e^{\vz_i}},
\end{equation}
where $\vp_{max} = \max_{i \neq y} \vp_{i}$. Compared to the classic margin loss defined on logits (see Table \ref{tab:1}), the logit values $\vz_{max/y}$ are substituted by the probability $\vp_{max/y}$ in $\Ls_{pm}$. Intuitively, probability-based loss functions take all logit values into consideration at their denominator, which opens up more attack possibilities toward different logit directions.

\begin{table*}[h]
  \centering
  \caption{A summary of the formulas and gradients of four popular adversarial loss functions and our probability margin loss. $\vz_{max}$/$p_{max}$ is the maximum value of $\vz_i$/$p_i$ for $i \neq y$, respectively.
  }
  \begin{adjustbox}{width=0.85\linewidth}
    \begin{tabular}{ccc|lll|lll}
    \toprule
    \multicolumn{3}{c|}{Adversarial Loss} & \multicolumn{3}{c|}{Loss Formula ($\Ls\left(\vz, y\right)$)} & \multicolumn{3}{c}{ Adversarial Gradient ($\triangledown_{\vx}\Ls\left(\vz,y\right)$)} \\
    \midrule
    \multicolumn{3}{c|}{\multirow{2}[1]{*}{\text{Untargeted CE ($\Ls_{ce}$)}}} & \multicolumn{3}{l|}{\multirow{2}[1]{*}{$-\log \vp_{y}= -\vz_y + \log\sum_{i=1}^N e^{\vz_i}$
    }} & \multicolumn{3}{l}{\multirow{2}[1]{*}{ $\sum_{i\neq y}^{N}\vp_{i}\triangledown_{\vx}\vz_{i}+\left(\vp_{y}-1\right)\triangledown_{\vx}\vz_y$
    }} \\
    \multicolumn{3}{c|}{} & \multicolumn{3}{l|}{} & \multicolumn{3}{l}{} \\ \hline
    \multicolumn{3}{c|}{\multirow{2}[0]{*}{Targeted CE ($\Ls_{cet}$)}} & \multicolumn{3}{l|}{\multirow{2}[0]{*}{ $\log \vp_{max}=\vz_{max} - \log{\sum_{i=1}^N e^{\vz_i}}$
    }} & \multicolumn{3}{l}{\multirow{2}[0]{*}{$-\sum_{i\neq max}^{N}\vp_{i}\triangledown_{\vx}\vz_{i}+\left(1-\vp_{max}\right)\triangledown_{\vx}\vz_{max}$
    }} \\
    \multicolumn{3}{c|}{} & \multicolumn{3}{l|}{} & \multicolumn{3}{l}{} \\ \hline
    \multicolumn{3}{c|}{\multirow{2}[0]{*}{DLR ($\Ls_{dlr}$)}} & \multicolumn{3}{l|}{\multirow{2}[0]{*}{$\frac{-\vz_y + \max_{i \neq y} \vz_i}{\vz_{\pi_1} - \vz_{\pi_3}}$
    }} & \multicolumn{3}{l}{\multirow{2}[0]{*}{
         $\frac{\triangledown_x\vz_{max}}{\vz_y-\vz_{\pi3}} + \frac{\left(\vz_{\pi3} -\vz_{max}\right)\triangledown_x\vz_{y} + \left(\vz_{max}-\vz_y\right)\triangledown_x\vz_{\pi3}}{\left(\vz_{\pi1}-\vz_{\pi3}\right)^2}$
    }} \\ 
    \multicolumn{3}{c|}{} & \multicolumn{3}{l|}{} & \multicolumn{3}{l}{} \\ \hline
    \multicolumn{3}{c|}{\multirow{2}[0]{*}{Margin ($\Ls_{mg}$)}} & \multicolumn{3}{l|}{\multirow{2}[0]{*}{$\vz_{max}-\vz_y$
    }} & \multicolumn{3}{l}{\multirow{2}[0]{*}{$\triangledown_{\vx}\vz_{max}-\triangledown_{\vx}\vz_y$
    }} \\ 
    \multicolumn{3}{c|}{} & \multicolumn{3}{l|}{} & \multicolumn{3}{l}{} \\ \hline
    \multicolumn{3}{c|}{\multirow{2}[1]{*}{Probability Margin ($\Ls_{pm}$)}} & \multicolumn{3}{l|}{\multirow{2}[1]{*}{$ \vp_{max} -\vp_y = \frac{e^{\vz_{max}}-e^{\vz_y}}{\sum_{i}^{N}e^{\vz_i}}$
    }} & \multicolumn{3}{l}{\multirow{2}[1]{*}{ $\left(\vp_y-\vp_{max}\right)\sum_{i}^{N}\vp_{i}\triangledown_{\vx}\vz_{i} + \vp_{max}\triangledown_{\vx}\vz_{max} - \vp_y\triangledown_{\vx}\vz_y
    $}} \\
    \multicolumn{3}{c|}{} & \multicolumn{3}{l|}{} & \multicolumn{3}{l}{} \\
    \bottomrule
    \end{tabular}%
    \end{adjustbox}
  \label{tab:1}%
\end{table*}%

\paragraph{Relationship to Existing Adversarial Losses}
We summarize the formulas and adversarial gradients of five adversarial loss functions: 1) untargeted cross-entropy (CE), 2) targeted cross-entropy, 3) DLR \cite{croce2020reliable}, 4) margin loss \cite{carlini2017towards}, and our 5) probability margin loss. 
The two targeted and untargeted CE use probabilities to compute the loss, while margin loss only considers the logits of two classes, and DLR exploits the logits of three classes. The DLR loss is more closely related to the margin loss but explores different targets than only the $\vz_{max}$. Our PM loss is also defined on probabilities which is similar to the two CE losses in this sense, but only on two classes which is similar to the margin loss. By incorporating different logits, our PM loss is also similar to DLR but in the exponential space. Another interesting observation is that \textbf{the margin loss is equivalent to the sum of the targeted CE loss and the untargeted CE loss}. This interesting observation implies that the effective margin loss verified in recent work \cite{ma2023imbalanced} can be obtained by combining the CE losses. Next, we will show from the gradient perspective that \textbf{our PM loss is a weighted combination of the targeted and untargeted CE losses} and thus is even more effective than the margin loss.

From the gradient formulas, it can be inferred that during the optimization process of untargeted CE, the value of \( \vz_y \) decreases while the values of \( \vz_{i \neq y } \) increase, with weights depending on the probability of the corresponding class. This provides multiple weighted adversarial directions for exploration but may have a problem in finding the optimal direction, as it is hard to concentrate on one particular direction. In contrast, targeted CE concentrates all exploration toward $\vz_{max}$ while suppressing all other possibilities via its first term. This resolves the previous issue of untargeted CE but hinders the exploration of alternative adversarial directions.
Margin loss increases $\vz_{max}$ while decreasing $\vz_y$, following the maximum adversarial direction, which is similar to targeted CE but without the probabilistic weighting. 
Compared to margin loss, DLR includes one more direction (i.e., $\pi_3$) to weight the $\vz_{max}$ but still omits other dimensions. In fact, the gradient of our PM loss is a weighted combination of the gradients of untargeted and targeted CE losses. The relationship can be derived as follows:
\begin{equation}
\begin{aligned}
    \triangledown_{\vx}\Ls_{pm}\left(\vz,y\right)
    &= \left(\vp_y-\vp_{max}\right)\sum_{i}^{N}\vp_{i}\triangledown_{\vx}\vz_i \\
    &+ \vp_{max}\triangledown_{\vx}\vz_{max} - \vp_y\triangledown_{\vx}\vz_y\\
    &=  \vp_y\triangledown_{\vx}\Ls_{ce}\left(\vz, y\right)+\vp_{max}\triangledown_{\vx}\Ls_{cet}\left(\vz, y\right).
\end{aligned}
\end{equation}
This means our PM loss enjoys the advantage of both untargeted and targeted loss, achieving a hybrid attack effectiveness with a single formula.
Moreover, the above weighted combination of $\Ls_{ce}$ and $\Ls_{cet}$ makes the PM loss more adaptive to targeted or untargeted attacks. Specifically, for all terms other than \( p_y \) and \( p_{\text{max}} \), there is a regularization effect that suppresses the increase of these terms by the difference \( p_y - p_{\text{max}} \). The closer \( p_{\text{max}} \) is to \( p_y \), the stronger the regularization effect becomes, 
allowing more focus on optimizing in the direction that maximizes \( z_{\text{max}} \).
\subsection{Probability Margin Attack (PMA)}
Our proposed PMA attack adopts the PM loss as the adversarial objective and follows the two-stage attack pipeline of the MD attack. The reason why we closely follow the MD pipeline as it already combines previous tricks such as multi-targeted attack, objective alternation, multi-stage exploration, random restart, and margin decomposition, into an integrated pipeline. By simply replacing the adversarial objective, we can easily improve the attack effectiveness with no additional costs.

The complete attack process of PMA involves alternating the attack objective between the two margin terms of the PM loss.
The extract loss term used in each stage of PMA is defined as follows:

\begin{gather}\label{eq}
    \vx_{k+1} = \prod_{\epsilon}(\vx_k + \alpha \cdot \text{sign}(\nabla_{\vx_k} \Ls_k^r(\vx_k, y)) \\
    \Ls_k^r(\vx_k, y) =
\begin{cases}
\vp_{max} & \text{if } k < K^{1} \quad\text{and}\quad r \% 2 = 0 \\
-\vp_y & \text{if } k < K^{1} \quad\text{and} \quad r \% 2 = 1 \\
\vp_{max} - \vp_y & \text{if } K^{1} < k < K.
\end{cases}
\end{gather}
In the above equations, $\prod$ denotes the projection operation, ensuring that the perturbations are within the $\epsilon$-ball centered at $\vx$. The attack step is represented by $k \in \{1, ..., K\}$, while $k < K^{1}$ denotes the attack stage 1 ($K^{1} \in [1,K]$). $r \in {1, ..., n}$ represents the $r$-th restart, which is used to calculate which loss term to use via the modulo operation (\%). The loss function $\Ls_k^r$ switches from the individual probability terms to the full PM loss at stage 2, which ensures the full strength of the attack. 
The corresponding adversarial gradients of the loss terms are as follows:
\begin{equation}\label{eq:pma}
\resizebox{1.0\hsize}{!}{
$\triangledown_{\vx}\mathcal{L}_k^r(\vx_k, y) =
\begin{cases}
\vp_{\text{max}}\left(-\sum_{i\neq \text{max}}^{N}\vp_{i}\triangledown_{\vx}\vz_{i}+\left(1-\vp_{\text{max}}\right)\triangledown_{\vx}\vz_{\text{max}}\right) & \text{if } k < K^{1} \text{ and } r \% 2 = 0 \\
\vp_y\left(\sum_{i\neq y}^{N} \vp_{i}\triangledown_{\vx}\vz_{i}+\left(\vp_{y}-1\right)\triangledown_{\vx}\vz_y\right) & \text{if } k < K^{1} \text{ and } r \%  2 = 1 \\
\left(\vp_y-\vp_{\text{max}}\right)\sum_{i}^{N}\vp_{i}\vz_{i} + \vp_{\text{max}}\triangledown_{\vx}\vz_{\text{max}} - \vp_y\triangledown_{\vx}\vz_y & \text{if } K^{1} < k < K.
\end{cases}$
}
\end{equation}
Our PMA ensures sufficient exploration of the $\vp_{\text{max}}$ and $\vp_{i \neq y}$ directions by either encouraging the $\vp_{\text{max}}$ gradient or depressing the $\vp_{y}$ gradient during the first stage ($k < K^1$). This is optimized alternatively along with multiple restarts. The perturbation obtained in stage 1 provides a good initialization for stage 2 which further attacks the full PM loss to find stronger attacks. The detailed procedure of our PMA is described in Algorithm \ref{alg:pma}.

\vspace{10pt}

\begin{algorithm}[h]
        \small
          \caption{Probability Margin Attack}
          \label{alg:pma}
        \begin{algorithmic}[1]
            \State {\bfseries Input:} clean sample $\bm{\vx}$, label $y$, model $f$, stage 1 steps $K^{1}$, total steps $K$
            \State {\bfseries Output:}  adversarial example $\bm{\vx}_{adv}$
            \State {\bfseries Parameters:} Maximum perturbation $\epsilon$, step size $\alpha$, number of restarts $n$, first stage steps $K^{1}$, total steps $K$
            \State $\bm{\vx}_{adv} \gets \bm{\vx}$
            \For{$ r \in \{1, ..., n\}$}
                \State $\bm{\vx}_0 \gets \bm{\vx} + uniform(-\epsilon,\epsilon)$ \Comment{uniform noise initialization}
                \For{$k \in \{1, ..., K\}$}
                \If{$k < K^{1}$}
                    \State $\alpha \gets \epsilon\cdot \big(1+\cos(\frac{k-1}{K^{1}}\pi)\big)$
                \ElsIf{$k \geq K^{1}$}
                    \State $\alpha \gets \epsilon\cdot \big(1+\cos(\frac{k-K^{1}}{K-K^{1}}\pi)\big)$
                \EndIf
                    \State $\bm{\vx}_k \gets \Pi_\epsilon \big(\bm{\vx}_{k-1} + \alpha\cdot\sign(\nabla_{\bm{\vx}} \Ls^r_k(\bm{\vx}_{k-1}, y))\big)$
                    \Comment{update $\bm{\vx}_k$ by \eqref{eq:pma}}
                    \If{$\Ls(\bm{\vx}_{adv}) < \Ls(\bm{\vx}_k)$}
                        \State $\bm{\vx}_{adv} \gets \bm{\vx}_k$
                    \EndIf
                \EndFor
            \EndFor
            \State $\bm{\vx}_{adv} = \Pi_{[0,1]}\big(\bm{\vx}_{adv}\big)$ \Comment{final clipping}
            \State {\bfseries return} $\bm{\vx}_{adv}$
        \end{algorithmic}
\end{algorithm}
\section{Experiments}
\label{sec:exp}

In this section, we first describe our experiment setup and then present the results for individual evaluation, ensemble evaluation, and large-scale evaluation, respectively.

\subsection{Experimental Setup}
\paragraph{Datasets and Models}
Following previous works, we use CIFAR-10, CIFAR-100, and ImageNet-1k as the evaluation datasets.
We choose adversarially trained models from the RobustBench leaderboard \cite{croce2021robustbench} as our target models, which are the most robust models to date. Specifically, we select the top 10 models from the CIFAR-10 leaderboard, the top 8 models from the CIFAR-100 leaderboard, and the top 11 models from the ImageNet leaderboard. These total of 29 models cover a variety of architectures, including ResNet\cite{he2016deep}, WideResNet\cite{zagoruyko2016wide}, ViT\cite{dosovitskiy2020image}, XCiT\cite{ali2021xcit}, Swin Transformer\cite{liu2021swin}, and ConvNeXt\cite{liu2022convnet}.

\begin{table*}[h]
  \centering
  \begin{adjustbox}{width=0.9\linewidth}
    \begin{tabular}{ll|c|ccccc|ccccc}
    \toprule
    \multicolumn{2}{c|}{\textbf{Defense Model}} & \textbf{Clean} & \textbf{$\text{PGD}_\text{ce}$} & \textbf{$\text{PGD}_\text{dlr}$} & \textbf{$\text{PGD}_\text{mg}$} & \textbf{$\text{PGD}_\text{pm}$} & \textbf{diff} & \textbf{$\text{APGD}_\text{ce}$} & \textbf{$\text{APGD}_\text{dlr}$} & \textbf{$\text{APGD}_\text{mg}$} & \textbf{$\text{APGD}_\text{pm}$} & \textbf{diff} \\
    \midrule
    \multicolumn{13}{c}{\textbf{CIFAR-10}} \\
    \midrule
    \multicolumn{2}{l|}{RWRN-70-16\cite{peng2023robust}} & 93.27 & 73.98 & 72.03 & 71.94 & \textbf{71.76} & \textbf{-0.18} & 73.84 & 72.02 & 71.94 & \textbf{71.78} & \textbf{-0.16} \\
    \multicolumn{2}{l|}{WRN-70-16\cite{wang2023better}} & 93.25 & 73.60 & 71.61 & 71.56 & \textbf{71.34} & \textbf{-0.22} & 73.45 & 71.60 & 71.54 & \textbf{71.39} & \textbf{-0.15} \\
    \multicolumn{2}{l|}{Mixing\cite{bai2023improving}} & 95.23 & 74.96 & 69.55 & 69.52 & \textbf{69.49} & \textbf{-0.03} & 74.59 & 69.37 & 69.28 & \textbf{69.22} & \textbf{-0.06} \\
    \multicolumn{2}{l|}{WRN-28-10\cite{cui2023decoupled}} & 92.16 & 70.65 & 68.65 & 68.62 & \textbf{68.47} & \textbf{-0.15} & 70.53 & 68.60 & 68.55 & \textbf{68.33} & \textbf{-0.22} \\
    \multicolumn{2}{l|}{WRN-28-10\cite{wang2023better}} & 92.44 & 70.31 & 68.31 & 68.22 & \textbf{68.10} & \textbf{-0.12} & 70.19 & 68.17 & 68.18 & \textbf{68.00} & \textbf{-0.17} \\
    \multicolumn{2}{l|}{WRN-70-16\cite{rebuffi2021fixing}} & 92.23 & 69.56 & 67.96 & 67.79 & \textbf{67.55} & \textbf{-0.24} & 69.40 & 67.86 & 67.70 & \textbf{67.39} & \textbf{-0.31} \\
    \multicolumn{2}{l|}{WRN-70-16\cite{gowal2021improving}} & 88.74 & 68.63 & 68.49 & 67.83 & \textbf{67.08} & \textbf{-0.75} & 68.48 & 68.28 & 67.68 & \textbf{67.03} & \textbf{-0.65} \\
    \multicolumn{2}{l|}{WRN-70-16\cite{gowal2020uncovering}} & 91.10 & 68.20 & 66.98 & 66.82 & \textbf{66.72} & \textbf{-0.10} & 68.04 & 66.85 & 66.72 & \textbf{66.55} & \textbf{-0.17} \\
    \multicolumn{2}{l|}{WRN-A4\cite{huang2023revisiting}} & 91.59 & 67.93 & 67.16 & 67.87 & \textbf{66.60} & \textbf{-0.56} & 67.80 & 67.05 & 66.82 & \textbf{66.59} & \textbf{-0.23} \\
    \multicolumn{2}{l|}{WRN-106-16\cite{rebuffi2021fixing}} & 88.50 & 67.64 & 65.69 & 65.65 & \textbf{65.46} & \textbf{-0.19} & 67.64 & 65.58 & 65.52 & \textbf{65.31} & \textbf{-0.21} \\
    \midrule
    \multicolumn{13}{c}{\textbf{CIFAR-100}} \\
    \midrule
    \multicolumn{2}{l|}{WRN-70-16\cite{wang2023better}} & 75.23 & 48.19 & 43.93 & 43.90 & \textbf{43.66} & \textbf{-0.24} & 48.14 & 43.84 & 43.84 & \textbf{43.53} & \textbf{-0.31} \\
    \multicolumn{2}{l|}{WRN-28-10\cite{cui2023decoupled}} & 73.83 & 43.83 & 40.34 & 40.33 & \textbf{40.13} & \textbf{-0.20} & 43.93 & 40.33 & 40.34 & \textbf{40.02} & \textbf{-0.31} \\
    \multicolumn{2}{l|}{WRN-28-10\cite{wang2023better}} & 72.58 & 44.09 & 39.69 & 39.62 & \textbf{39.47} & \textbf{-0.15} & 44.09 & 39.60 & 39.63 & \textbf{39.34} & \textbf{-0.26} \\
    \multicolumn{2}{l|}{WRN-70-16\cite{gowal2020uncovering}} & 69.15 & 40.04 & 39.11 & 38.93 & \textbf{38.05} & \textbf{-0.12} & 39.94 & 38.95 & 38.76 & \textbf{37.88} & \textbf{-0.88} \\
    \multicolumn{2}{l|}{XCiT-L12\cite{debenedetti2023light}} & 70.77 & 38.98 & 37.20 & 36.72 & \textbf{36.03} & \textbf{-0.31} & 38.93 & 37.21 & 36.74 & \textbf{35.95} & \textbf{-0.79} \\
    \multicolumn{2}{l|}{WRN-70-16\cite{rebuffi2021fixing}} & 63.56 & 38.39 & 36.20 & 36.19 & \textbf{35.61} & \textbf{-0.58} & 38.24 & 36.09 & 36.08 & \textbf{35.47} & \textbf{-0.61} \\
    \multicolumn{2}{l|}{XCiT-M12\cite{debenedetti2023light}} & 69.20 & 38.87 & 36.15 & 35.80 & \textbf{35.16} & \textbf{-0.64} & 38.73 & 36.00 & 35.68 & \textbf{35.05} & \textbf{-0.63} \\
    \multicolumn{2}{l|}{WRN-70-16\cite{pang2022robustness}} & 65.56 & 36.73 & 34.23 & 34.16 & \textbf{33.82} & \textbf{-0.34} & 36.64 & 34.19 & 34.15 & \textbf{33.77} & \textbf{-0.38} \\
    \midrule
    \multicolumn{13}{c}{\textbf{ImageNet-1k}} \\
    \midrule
    \multicolumn{2}{l|}{Swin-L\cite{liu2023comprehensive}} & 78.18 & 59.47 & 59.80 & 59.31 & \textbf{57.97} & \textbf{-1.34} & 59.25 & 59.69 & 59.63 & \textbf{57.85} & \textbf{-1.40} \\
    \multicolumn{2}{l|}{Mixing\cite{bai2024mixednuts}} & 81.10 & 67.55  &  61.05 & 59.79  &   \textbf{59.20}  & \textbf{-0.59} & 67.24 & 60.94 & 59.56 & \textbf{59.18} & \textbf{-0.38} \\
    \multicolumn{2}{l|}{ConvNeXt-L\cite{liu2023comprehensive}} & 77.48 & 58.42 & 59.17 & 58.63 & \textbf{57.24} & \textbf{-1.18} & 58.15 & 58.15 & 58.59 & \textbf{57.19} & \textbf{-0.96} \\
    \multicolumn{2}{l|}{ConvNeXt-L+CS\cite{singh2024revisiting}} & 76.79 & 57.87 & 58.56 & 57.88 & \textbf{56.50} & \textbf{-1.37} & 57.63 & 58.43 & 57.82 & \textbf{56.47} & \textbf{-1.16} \\
    \multicolumn{2}{l|}{Swin-B\cite{liu2023comprehensive}} & 76.22 & 57.26 & 56.69 & 56.28 & \textbf{54.93} & \textbf{-1.35} & 56.96 & 56.65 & 56.25 & \textbf{54.91} & \textbf{-1.34} \\
    \multicolumn{2}{l|}{ConvNeXt-B+CS\cite{singh2024revisiting}} & 75.46 & 56.10 & 56.45 & 55.91 & \textbf{54.55} & \textbf{-1.36} & 55.79 & 56.36 & 55.86 & \textbf{54.52} & \textbf{-1.27} \\
    \multicolumn{2}{l|}{ConvNeXt-B\cite{liu2023comprehensive}} & 76.38 & 56.37 & 56.86 & 56.20 & \textbf{54.77} & \textbf{-1.43} & 56.04 & 56.78 & 56.19 & \textbf{54.74} & \textbf{-1.30} \\
    \multicolumn{2}{l|}{ViT-B+CS\cite{singh2024revisiting}} & 76.12 & 55.34 & 55.52 & 55.01 & \textbf{53.61} & \textbf{-1.40} & 55.05 & 55.40 & 54.94 & \textbf{53.55} & \textbf{-1.39} \\
    \multicolumn{2}{l|}{ConvNeXt-S+CS\cite{singh2024revisiting}} & 73.37 & 52.69 & 52.72 & 51.94 & \textbf{50.38} & \textbf{-1.56} & 52.32 & 52.58 & 51.91 & \textbf{50.34} & \textbf{-1.56} \\
    \multicolumn{2}{l|}{ConvNeXt-T+CS\cite{singh2024revisiting}} & 72.45 & 51.20 & 50.29 & 49.80 & \textbf{48.28} & \textbf{-1.52} & 50.79 & 50.24 & 49.78 & \textbf{48.23} & \textbf{-1.56} \\
    \multicolumn{2}{l|}{RWRN-101-2\cite{peng2023robust}} & 73.45 & 51.02 & 51.65 & 51.22 & \textbf{49.67} & \textbf{-1.35} & 50.76 & 51.57 & 51.15 & \textbf{49.62} & \textbf{-1.14} \\
    \bottomrule
    \end{tabular}%
    \end{adjustbox}
  \caption{The robust accuracy (\%) of different models. The subscripts of the attacks (columns) represent the adversarial loss function: cross-entropy (ce), DLR loss (dlr), margin loss (mg), and probability margin loss (pm). The “diff” column shows the decrease in robust accuracy when evaluated with our pm loss compared to the best baseline loss function. The best results are highlighted in bold.}
  \label{tab:2}%
\end{table*}%

\paragraph{Attack Setting and Baselines}

\vspace{-15pt}

Following previous works~\cite{croce2021robustbench, croce2020reliable}, we focus on $L_{\infty}$ norm adversarial attack and set the perturbation budget to $\epsilon = 8/255$ for CIFAR-10 and CIFAR-100, $\epsilon = 4/255$ for ImageNet-1k. For untargeted attacks, the number of attack steps is set to 100, while for targeted (or multi-targeted) attacks, the step for each target is set to 100.
For untargeted attacks, we compare our PM loss to several existing loss functions, including cross-entropy (CE), DLR \cite{croce2020reliable}, margin loss \cite{carlini2017towards}, mixed loss function~\cite{antoniou2022alternating}, and MIFPE\cite{yu2023efficient}, combined with three attack strategies: the classic PGD strategy \cite{madry2017towards}, the APGD strategy\cite{croce2020reliable}, and the Margin Decomposition strategy\cite{ma2023imbalanced}. Notably, for the Margin Decomposition strategy, we set the hyperparameter $K'$ to 25 and the number of restarts to 1. The effects of varying $K'$ values and the number of restarts $n$ are further analyzed in our ablation study, which is detailed in Appendix B.
FAB \cite{croce2020minimally} is also evaluated.
For targeted attacks, we evaluate the DLR, margin loss, and our probability margin loss with APGDT\cite{croce2020reliable} attack strategy. This evaluation includes a total of 9 targets, with each target subjected to 100 attack steps. We report the robust accuracies of different defense models evaluated under these attacks \cite{croce2020reliable}.

In Appendix C, we present comparative experiments involving the ACG\cite{yamamura2022diversified} and AAA\cite{liu2022practical} methods. Furthermore, Appendix D provides a comparison of experimental results between the traditional SGD+sign update strategy and the optimization-based update strategy.

We report the robust accuracies of different defense models evaluated under these attacks \cite{croce2020reliable}.

\subsection{Main Results}

\paragraph{Effectiveness of the PM Loss}
We first fix the attack strategy to that of PGD and APGD, and then compare our proposed PM loss with other existing loss functions. As shown in Table \ref{tab:2}, our PM loss outperforms CE, DLR, and margin losses across all evaluated defense models. The robustness evaluated using the PM loss is about 0.03\% to 1.56\% lower than that evaluated using the existing loss functions.
The advantage of the PM loss is more pronounced on datasets with more classes.
Particularly, on ImageNet-1k, our PM loss achieves an average robustness reduction of 1.22\%, demonstrating its potential for large-scale evaluation.
Given the PM loss's composition of $\vp_{max}$ and $\vp_{y}$ with variable weighting, we assessed its sensitivity to these weights via an ablation study, with results detailed in Appendix E.

\begin{table*}[htbp]
  \centering
  
  \begin{adjustbox}{width=0.9\linewidth}
    \setlength{\tabcolsep}{4pt}
    \begin{tabular}{ll|c|cccccccccc}
    \toprule
    \multicolumn{2}{c|}{\textbf{Defense}} &\textbf{Clean} & \textbf{$\text{PGD}_\text{pm}$} & \textbf{$\text{PGD}_\text{alt}$} & \textbf{$\text{PGD}_\text{mi}$} & \textbf{$\text{APGD}_\text{pm}$} & \textbf{$\text{APGD}_\text{mi}$} & \textbf{$\text{FAB}$} & \textbf{$\text{MD}$} & \textbf{$\text{Ours}$} & \textbf{diff} & \textbf{$\text{AA}$} \\
    \midrule
    \multicolumn{13}{c}{\textbf{CIFAR-10}} \\
    \midrule
    \multicolumn{2}{l|}{RWRN-70-16\cite{peng2023robust}} & 93.27 & 71.76 & 71.15 & 71.25 & 71.78 & 71.30 & 72.31 & 71.14 & \textbf{71.10} & \textbf{-0.05} & 71.10 \\
    \multicolumn{2}{l|}{WRN-70-16\cite{wang2023better}} & 93.25 & 71.34 & 70.87 & 70.82 & 71.39 & 70.92 & 71.90 & 70.74 & \textbf{70.67} & \textbf{-0.07} & 70.70 \\
    \multicolumn{2}{l|}{Mixing\cite{bai2023improving}} & 95.23 & 69.49 & 68.98 & 68.96 & 69.22 & 70.43 & 70.23 & 68.59 & \textbf{68.43} & \textbf{-0.16} & 68.06 \\
    \multicolumn{2}{l|}{WRN-28-10\cite{cui2023decoupled}} & 92.16 & 68.47 & 67.88 & 67.95 & 68.33 & 68.03 & 68.99 & 67.79 & \textbf{67.72} & \textbf{-0.07} & 67.75 \\
    \multicolumn{2}{l|}{WRN-28-10\cite{wang2023better}} & 92.44 & 68.10 & 67.46 & 67.51 & 68.00 & 67.56 & 68.62 & 67.42 & \textbf{67.33} & \textbf{-0.09} & 67.31 \\
    \multicolumn{2}{l|}{WRN-70-16\cite{rebuffi2021fixing}} & 92.23 & 67.55 & 66.94 & 66.86 & 67.39 & 67.37 & 67.80 & 66.84 & \textbf{66.80} & \textbf{-0.04} & 66.59 \\
    \multicolumn{2}{l|}{WRN-70-16\cite{gowal2021improving}} & 88.74 & 67.08 & 66.35 & 66.41 & 67.03 & 66.77 & 67.33 & 66.68 & \textbf{66.24} & \textbf{-0.04} & 66.14 \\
    \multicolumn{2}{l|}{WRN-70-16\cite{gowal2020uncovering}} & 91.10 & 66.72 & 66.03 & 66.09 & 66.55 & 66.34 & 67.14 & 66.04 & \textbf{65.95} & \textbf{-0.09} & 65.89 \\
    \multicolumn{2}{l|}{WRN-A4\cite{huang2023revisiting}} & 91.59 & 66.60 & 65.98 & 66.02 & 66.59 & 66.09 & 67.62 & 65.90 & \textbf{65.87} & \textbf{-0.03} & 65.78 \\
    \multicolumn{2}{l|}{WRN-106-16\cite{rebuffi2021fixing}} & 88.50 & 65.46 & 64.85 & 64.85 & 65.31 & 65.18 & 65.62 & 64.84 & \textbf{64.69} & \textbf{-0.15} & 64.68 \\
    \midrule
    \multicolumn{13}{c}{\textbf{CIFAR-100}} \\
    \midrule
    \multicolumn{2}{l|}{WRN-70-16\cite{wang2023better}} & 75.23 & 43.66 & 43.00 & 43.03 & 43.53 & 43.11 & 43.89 & 42.86 & \textbf{42.83} & \textbf{-0.03} & 42.68 \\
    \multicolumn{2}{l|}{WRN-28-10\cite{cui2023decoupled}} & 73.83 & 40.13 & 39.52 & 39.51 & 40.02 & 39.56 & 40.69 & 39.46 & \textbf{39.39} & \textbf{-0.07} & 39.20 \\
    \multicolumn{2}{l|}{WRN-28-10\cite{wang2023better}} & 72.58 & 39.47 & 39.07 & 39.05 & 39.34 & 39.15 & 40.01 & 38.93 & \textbf{38.92} & \textbf{-0.10} & 38.79 \\
    \multicolumn{2}{l|}{WRN-70-16\cite{gowal2020uncovering}} & 69.15 & 38.05 & 37.38 & 37.38 & 37.88 & 37.61 & 37.96 & 37.31 & \textbf{37.20} & \textbf{-0.11} & 36.89 \\
    \multicolumn{2}{l|}{XCiT-L12\cite{debenedetti2023light}} & 70.77 & 36.03 & 35.44 & 35.42 & 35.95 & 35.40 & 36.07 & 35.34 & \textbf{35.27} & \textbf{-0.07} & 35.04 \\
    \multicolumn{2}{l|}{WRN-70-16\cite{rebuffi2021fixing}} & 63.56 & 35.61 & 34.90 & 34.84 & 35.47 & 34.96 & 35.62 & 34.79 & \textbf{34.74} & \textbf{-0.05} & 34.68 \\
    \multicolumn{2}{l|}{XCiT-M12\cite{debenedetti2023light}} & 69.20 & 35.16 & 34.48 & 34.43 & 35.05 & 34.40 & 34.97 & 34.48 & \textbf{34.33} & \textbf{-0.10} & 34.20 \\
    \multicolumn{2}{l|}{WRN-70-16\cite{pang2022robustness}} & 65.56 & 33.82 & 33.32 & 33.32 & 33.77 & 33.33 & 33.70  & 33.25 & \textbf{33.14} & \textbf{-0.11} & 33.04 \\
    \midrule
    \multicolumn{13}{c}{\textbf{ImageNet-1k}} \\
    \midrule
    \multicolumn{2}{l|}{Swin-L\cite{liu2023comprehensive}} & 78.18 & 57.97 & 57.50 & 57.58 & 57.85 & 57.55 & -    & 57.42 & \textbf{57.35} & \textbf{-0.07} & 57.26 \\
    \multicolumn{2}{l|}{Mixing\cite{bai2024mixednuts}} & 81.10 &  59.20 & 59.29 & 62.53 & 59.18 & 62.36 & -    & 58.69 & \textbf{58.65} & \textbf{-0.04} & 58.31 \\
    \multicolumn{2}{l|}{ConvNeXt-L\cite{liu2023comprehensive}} & 77.48 & 57.24 & 56.63 & 56.69 & 57.19 & 56.69 & -    & 56.64 & \textbf{56.53} & \textbf{-0.10} & 56.42 \\
    \multicolumn{2}{l|}{ConvNeXt-L+CS\cite{singh2024revisiting}} & 76.79 & 56.50 & 56.08 & 56.20 & 56.47 & 56.19 & -    & 56.49 & \textbf{55.94} & \textbf{-0.14} & 55.86 \\
    \multicolumn{2}{l|}{Swin-B\cite{liu2023comprehensive}} & 76.22 & 54.93 & 54.55 & 54.57 & 54.91 & 54.57 & -    & 54.48 & \textbf{54.41} & \textbf{-0.07} & 54.30 \\
    \multicolumn{2}{l|}{ConvNeXt-B+CS\cite{singh2024revisiting}} & 75.46 & 54.55 & 54.08 & 54.15 & 54.52 & 54.14 & -    & 54.39 & \textbf{53.94} & \textbf{-0.20} & 53.84 \\
    \multicolumn{2}{l|}{ConvNeXt-B\cite{liu2023comprehensive}} & 76.38 & 54.77 & 54.23 & 54.27 & 54.74 & 54.27 & -   & 54.28 & \textbf{54.13} & \textbf{-0.14} & 54.04 \\
    \multicolumn{2}{l|}{ViT-B+CS\cite{singh2024revisiting}} & 76.12 & 53.61 & 53.00 & 53.19 & 53.55 & 53.14 & -   & 53.34 & \textbf{52.82} & \textbf{-0.18} & 52.66 \\
    \multicolumn{2}{l|}{ConvNeXt-S+CS\cite{singh2024revisiting}} & 73.37 & 50.38 & 49.93 & 49.98 & 50.34 & 49.96 & - & 50.12 & \textbf{49.74} & \textbf{-0.22} & 49.65 \\
    \multicolumn{2}{l|}{ConvNeXt-T+CS\cite{singh2024revisiting}} & 72.45 & 48.28 & 47.89 & 47.95 & 48.23 & 47.93 & -  & 48.18 & \textbf{47.70} & \textbf{-0.19} & 47.60 \\
    \multicolumn{2}{l|}{RWRN-101-2\cite{peng2023robust}} & 73.45 & 49.67 & 49.19 & 49.29 & 49.62 & 49.30 & - & 49.25 & \textbf{49.06} & \textbf{-0.13} & 48.96 \\
    \bottomrule
    \end{tabular}
    \end{adjustbox}
  \caption{The models' robustness (\%) evaluated by individual attacks. The "diff" column marks the robustness decrease by our PMA compared to the best baseline. The best results are boldfaced.}
  \label{tab:3}%
\end{table*}%

\vspace{-10pt}

\paragraph{Effectiveness of PMA}
We then compare the effectiveness of our PMA with other existing attack methods. As shown in Table \ref{tab:3}, the robustness obtained by our PMA is the lowest. Particularly, PMA reduces the robustness of different models evaluated by other attacks by up to 0.22\%.

\begin{table*}[h]
  \centering
  \begin{adjustbox}{width=0.9\linewidth}
    \begin{tabular}{|ll|r|rr|rr|rr|rr|rr|rr|rr|rr|}
    \toprule
    \multicolumn{2}{|c|}{\textbf{Defense}} & \multicolumn{1}{l|}{\textbf{PMA}} & \multicolumn{2}{c|}{$\textbf{PGD}_\textbf{pm}$} & \multicolumn{2}{c|}{$\textbf{APGD}_\textbf{pm}$} & \multicolumn{2}{c|}{$\textbf{APGDT}_\textbf{pm}$} & \multicolumn{2}{c|}{\textbf{MD}} & \multicolumn{2}{c|}{$\textbf{PGD}_\textbf{ce}$} & \multicolumn{2}{c|}{$\textbf{PGD}_\textbf{dlr}$} & \multicolumn{2}{c|}{$\textbf{PGD}_\textbf{mg}$} & \multicolumn{2}{c|}{$\textbf{PGD}_\textbf{mi}$} \\
    \midrule
    \multicolumn{2}{|l|}{RWRN-70-16\cite{peng2023robust}} & 71.10  & 71.06 & \textbf{-0.04 } & 71.08 & \textbf{-0.02 } & 71.05 & \textbf{-0.05 } & 71.07 & \textbf{-0.03 } & 71.10 & \textbf{0.00} & 71.06 & \textbf{-0.04 } & 71.08 & \textbf{-0.02 } & 71.08 & \textbf{-0.02 } \\
    \multicolumn{2}{|l|}{WRN-70-16\cite{wang2023better}} & 70.67  & 70.67 & \textbf{0.00 } & 70.66 & \textbf{-0.01 } & 70.60 & \textbf{-0.07 } & 70.64 & \textbf{-0.03 } & 70.67 & \textbf{0.00} & 70.67 & \textbf{0.00 } & 70.66 & \textbf{-0.01 } & 70.63 & \textbf{-0.04 } \\
    \multicolumn{2}{|l|}{Mixing\cite{bai2023improving}} & 68.43  & 68.05 & \textbf{-0.38 } & 68.10 & \textbf{-0.33 } & 67.93 & \textbf{-0.50 } & 68.26 & \textbf{-0.17 } & 68.42 & \textbf{-0.01} & 68.04 & \textbf{-0.39 } & 68.09 & \textbf{-0.34 } & 68.08 & \textbf{-0.35 } \\
    \multicolumn{2}{|l|}{WRN-28-10\cite{cui2023decoupled}} & 67.72  & 67.71 & \textbf{-0.01 } & 67.71 & \textbf{-0.01 } & 67.64 & \textbf{-0.08 } & 67.66 & \textbf{-0.06 } & 67.72 & \textbf{0.00} & 67.72 & \textbf{0.00 } & 67.71 & \textbf{-0.01 } & 67.72 & \textbf{0.00 } \\
    \multicolumn{2}{|l|}{WRN-28-10\cite{wang2023better}} & 67.32  & 67.30 & \textbf{-0.02 } & 67.31 & \textbf{-0.01 } & 67.26 & \textbf{-0.06 } & 67.29 & \textbf{-0.03 } & 67.32 & \textbf{0.00} & 67.31 & \textbf{-0.01 } & 67.31 & \textbf{-0.01 } & 67.30 & \textbf{-0.02 } \\
    \multicolumn{2}{|l|}{WRN-70-16\cite{rebuffi2021fixing}} & 66.80  & 66.75 & \textbf{-0.05 } & 66.72 & \textbf{-0.08 } & 66.61 & \textbf{-0.19 } & 66.72 & \textbf{-0.08 } & 66.77 & \textbf{-0.03} & 66.72 & \textbf{-0.08 } & 66.75 & \textbf{-0.05 } & 66.70 & \textbf{-0.10 } \\
    \multicolumn{2}{|l|}{WRN-70-16\cite{gowal2021improving}} & 66.24  & 66.19 & \textbf{-0.05 } & 66.18 & \textbf{-0.06 } & 66.13 & \textbf{-0.11 } & 66.16 & \textbf{-0.08 } & 66.23 & \textbf{-0.01} & 66.21 & \textbf{-0.03 } & 66.20 & \textbf{-0.04 } & 66.17 & \textbf{-0.07 } \\
    \multicolumn{2}{|l|}{WRN-70-16\cite{gowal2020uncovering}} & 65.95  & 65.94 & \textbf{-0.01 } & 65.92 & \textbf{-0.03 } & 65.86 & \textbf{-0.09 } & 65.92 & \textbf{-0.03 } & 65.95 & \textbf{0.00} & 65.92 & \textbf{-0.03 } & 65.94 & \textbf{-0.01 } & 65.92 & \textbf{-0.03 } \\
    \multicolumn{2}{|l|}{WRN-A4\cite{huang2023revisiting}} & 65.87  & 65.86 & \textbf{-0.01 } & 65.84 & \textbf{-0.03 } & 65.70  & \textbf{-0.17 } & 65.79 & \textbf{-0.08 } & 65.86 & \textbf{-0.01} & 65.86 & \textbf{-0.01 } & 65.87 & \textbf{0.00 } & 65.77 & \textbf{-0.10 } \\
    \multicolumn{2}{|l|}{WRN-106-16\cite{rebuffi2021fixing}} & 64.69  & 64.65 & \textbf{-0.04 } & 64.65 & \textbf{-0.04 } & 64.58  & \textbf{-0.11 } & 64.67 & \textbf{-0.02 } & 64.69 & \textbf{0.00} & 64.66 & \textbf{-0.03 } & 64.68 & \textbf{-0.01 } & 64.66 & \textbf{-0.03 } \\
    \midrule
    \multicolumn{18}{c}{}                                                                                                                         & \multicolumn{1}{c}{} \\
    \midrule
    \multicolumn{2}{|c|}{\textbf{Defense}} & \multicolumn{1}{l|}{\textbf{PMA}} & \multicolumn{2}{c|}{$\textbf{PGD}_\textbf{alt}$} & \multicolumn{2}{c|}{$\textbf{APGD}_\textbf{ce}$} & \multicolumn{2}{c|}{$\textbf{APGD}_\textbf{dlr}$} & \multicolumn{2}{c|}{$\textbf{APGD}_\textbf{mg}$} & \multicolumn{2}{c|}{$\textbf{APGD}_\textbf{mi}$} & \multicolumn{2}{c|}{$\textbf{APGDT}_\textbf{dlr}$} & \multicolumn{2}{c|}{$\textbf{APGD}_\textbf{mg}$} & \multicolumn{2}{c|}{\textbf{FAB}} \\
    \midrule
    \multicolumn{2}{|l|}{RWRN-70-16\cite{peng2023robust}} & 71.10  & 71.09 & \textbf{-0.01 } & 71.09 & \textbf{-0.01 } & 71.06 & \textbf{-0.04 } & 71.06 & \textbf{-0.04 } & 71.08 & \textbf{-0.02 } & 71.05 & \textbf{-0.05 } & 71.05 & \textbf{-0.05 } & 71.07 & \textbf{-0.03 } \\
    \multicolumn{2}{|l|}{WRN-70-16\cite{wang2023better}} & 70.67  & 70.65 & \textbf{-0.02 } & 70.67 & \textbf{0.00 } & 70.67 & \textbf{0.00 } & 70.66 & \textbf{-0.01 } & 70.64 & \textbf{-0.03 } & 70.61 & \textbf{-0.06 } & 70.61 & \textbf{-0.06 } & 70.65 & \textbf{-0.02 } \\
    \multicolumn{2}{|l|}{Mixing\cite{bai2023improving}} & 68.43  & 67.96 & \textbf{-0.47 } & 68.40 & \textbf{-0.03 } & 68.12 & \textbf{-0.31 } & 68.13 & \textbf{-0.30 } & 68.38 & \textbf{-0.05 } & 67.91 & \textbf{-0.52 } & 67.91 & \textbf{-0.52 } & 68.16 & \textbf{-0.27 } \\
    \multicolumn{2}{|l|}{WRN-28-10\cite{cui2023decoupled}} & 67.72  & 67.71 & \textbf{-0.01 } & 67.72 & \textbf{0.00 } & 67.69 & \textbf{-0.03 } & 67.70 & \textbf{-0.02 } & 67.71 & \textbf{-0.01 } & 67.63 & \textbf{-0.09 } & 67.64 & \textbf{-0.08 } & 67.72 & \textbf{0.00 } \\
    \multicolumn{2}{|l|}{WRN-28-10\cite{wang2023better}} & 67.32  & 67.31 & \textbf{-0.01 } & 67.32 & \textbf{0.00 } & 67.30 & \textbf{-0.02 } & 67.31 & \textbf{-0.01 } & 67.29 & \textbf{-0.03 } & 67.20 & \textbf{-0.12 } & 67.25 & \textbf{-0.07 } & 67.29 & \textbf{-0.03 } \\
    \multicolumn{2}{|l|}{WRN-70-16\cite{rebuffi2021fixing}} & 66.80  & 66.73 & \textbf{-0.07 } & 66.77 & \textbf{-0.03 } & 66.74 & \textbf{-0.06 } & 66.74 & \textbf{-0.06 } & 66.73 & \textbf{-0.07 } & 66.57 & \textbf{-0.23 } & 66.59 & \textbf{-0.21 } & 66.71 & \textbf{-0.09 } \\
    \multicolumn{2}{|l|}{WRN-70-16\cite{gowal2021improving}} & 66.24  & 66.16 & \textbf{-0.08 } & 66.21 & \textbf{-0.03 } & 66.20 & \textbf{-0.04 } & 66.20 & \textbf{-0.04 } & 66.19 & \textbf{-0.05 } & 66.09 & \textbf{-0.15 } & 66.13 & \textbf{-0.11 } & 66.17 & \textbf{-0.07 } \\
    \multicolumn{2}{|l|}{WRN-70-16\cite{gowal2020uncovering}} & 65.95  & 65.90 & \textbf{-0.05 } & 65.95 & \textbf{0.00 } & 65.89 & \textbf{-0.06 } & 65.91 & \textbf{-0.04 } & 65.95 & \textbf{0.00 } & 65.86 & \textbf{-0.09 } & 65.86 & \textbf{-0.09 } & 65.93 & \textbf{-0.02 } \\
    \multicolumn{2}{|l|}{WRN-A4\cite{huang2023revisiting}} & 65.87  & 65.82 & \textbf{-0.05 } & 65.87 & \textbf{0.00 } & 65.87 & \textbf{0.00 } & 65.86 & \textbf{-0.01 } & 65.82 & \textbf{-0.05 } & 65.73 & \textbf{-0.14 } & 65.72 & \textbf{-0.15 } & 65.80 & \textbf{-0.07 } \\
    \multicolumn{2}{|l|}{WRN-106-16\cite{rebuffi2021fixing}} & 64.69  & 64.66 & \textbf{-0.03 } & 64.69 & \textbf{0.00 } & 64.64 & \textbf{-0.05 } & 64.64 & \textbf{-0.05 } & 64.67 & \textbf{-0.02 } & 64.61 & \textbf{-0.08 } & 64.60 & \textbf{-0.09 } & 64.68 & \textbf{-0.01 } \\
    \bottomrule
    \end{tabular}%
    \end{adjustbox}
    \caption{The CIFAR-10 models' robustness (\%) evaluated by our PMA plus
  one existing attack (each column is a unique combination). The left and right numbers within the column show the absolute value and robustness decreases, respectively.
  }
  \label{tab:PMA+1}%
\end{table*}%

Compared with the recent MD attack which has the same attack pipeline as our PMA, the model robustness tested by PMA is reduced by up to 0.55\%. These results confirm that our PMA is the strongest individual attack in the current literature.
In the last column of Table \ref{tab:3}, we list the robustness evaluated by an ensemble attack: Auto Attack(AA). It can be observed that the results of PMA are very close to that of AA, which requires 4900 attack steps to achieve this strength. Our attack even surpasses the AA on three models on CIFAR-10, however, its running time is only 3\% of AA.
For a detailed comparison of the running times, please refer to the additional experiments provided in Appendix F.

\subsection{PMA Ensemble}
Here, we explore ensemble attacks with PMA. Specifically, we explore two ensemble strategies: 1) \textbf{\emph{PMA+1 ensemble}} that combines our PMA with one more other attack, and 2) \textbf{\emph{cascade ensemble}} that incorporates the remaining best attack in a sequence order until all popular individual attacks are included. As shown in Table \ref{tab:PMA+1}, the \textbf{\emph{PMA+1 ensemble}} always leads to better attacks that further reduce model robustness.
Among all the combinations, PMA + APGDT (multi-targeted APGD attack) produces the best performance, which reduces model robustness by 0.52\% in the best case. This implies that PMA and AGPDT may explore the vulnerabilities of a model in the most distinct directions.

\begin{table*}[h]
  \centering
  \begin{adjustbox}{width=0.9\linewidth}
    \begin{tabular}{|ll|r|rr|rr|rr|rr|rr|rr|rr|rr|r|}
    \toprule
    \multicolumn{2}{|c|}{\textbf{Defense}} & \multicolumn{1}{c|}{\textbf{AA}} & \multicolumn{1}{l|}{\textbf{PMA}} & \multicolumn{2}{c|}{\textbf{+$\text{PGD}_\text{pm}$}} & \multicolumn{2}{c|}{\textbf{+$\text{APGD}_\text{pm}$}} & \multicolumn{2}{c|}{\textbf{+$\text{APGDT}_\text{pm}$}} & \multicolumn{2}{c|}{\textbf{+MD}} & \multicolumn{2}{c|}{\textbf{+$\text{PGD}_\text{ce}$}} & \multicolumn{2}{c|}{\textbf{+$\text{PGD}_\text{dlr}$}} & \multicolumn{2}{c|}{\textbf{+$\text{PGD}_\text{mg}$}} & \multicolumn{2}{c|}{\textbf{$+\text{PGD}_\text{mi}$}} \\
    \midrule
    \multicolumn{2}{|l|}{RWRN-70-16\cite{peng2023robust}} & 71.10 & \multicolumn{1}{r|}{71.10 } & \multicolumn{1}{r}{71.06 } & \multicolumn{1}{r|}{\textbf{-0.04 }} & \multicolumn{1}{r}{71.06 } & \multicolumn{1}{r|}{\textbf{0.00 }} & \multicolumn{1}{r}{71.03 } & \multicolumn{1}{r|}{\textbf{-0.03 }} & \multicolumn{1}{r}{71.01 } & \multicolumn{1}{r|}{\textbf{-0.02 }} & \multicolumn{1}{r}{71.01 } & \multicolumn{1}{r|}{\textbf{0.00 }} & \multicolumn{1}{r}{71.00 } & \multicolumn{1}{r|}{\textbf{-0.01 }} & \multicolumn{1}{r}{71.00 } & \multicolumn{1}{r|}{\textbf{0.00 }} & \multicolumn{1}{r}{71.00 } & \textbf{0.00 } \\
    \multicolumn{2}{|l|}{WRN-70-16\cite{wang2023better}} & 70.70 & \multicolumn{1}{r|}{70.67 } & \multicolumn{1}{r}{70.67 } & \multicolumn{1}{r|}{\textbf{0.00 }} & \multicolumn{1}{r}{70.66 } & \multicolumn{1}{r|}{\textbf{-0.01 }} & \multicolumn{1}{r}{70.60 } & \multicolumn{1}{r|}{\textbf{-0.06 }} & \multicolumn{1}{r}{70.59 } & \multicolumn{1}{r|}{\textbf{-0.01 }} & \multicolumn{1}{r}{70.59 } & \multicolumn{1}{r|}{\textbf{0.00 }} & \multicolumn{1}{r}{70.59 } & \multicolumn{1}{r|}{\textbf{0.00 }} & \multicolumn{1}{r}{70.59 } & \multicolumn{1}{r|}{\textbf{0.00 }} & \multicolumn{1}{r}{70.57 } & \textbf{-0.02 } \\
    \multicolumn{2}{|l|}{Mixing\cite{bai2023improving}} & 68.06 & \multicolumn{1}{r|}{68.43 } & \multicolumn{1}{r}{68.05 } & \multicolumn{1}{r|}{\textbf{-0.38 }} & \multicolumn{1}{r}{67.86 } & \multicolumn{1}{r|}{\textbf{-0.19 }} & \multicolumn{1}{r}{67.64 } & \multicolumn{1}{r|}{\textbf{-0.22 }} & \multicolumn{1}{r}{67.58 } & \multicolumn{1}{r|}{\textbf{-0.06 }} & \multicolumn{1}{r}{67.58 } & \multicolumn{1}{r|}{\textbf{0.00 }} & \multicolumn{1}{r}{67.47 } & \multicolumn{1}{r|}{\textbf{-0.11 }} & \multicolumn{1}{r}{67.43 } & \multicolumn{1}{r|}{\textbf{-0.04 }} & \multicolumn{1}{r}{67.37 } & \textbf{-0.06 } \\
    \multicolumn{2}{|l|}{WRN-28-10\cite{cui2023decoupled}} & 67.75 & \multicolumn{1}{r|}{67.72 } & \multicolumn{1}{r}{67.71 } & \multicolumn{1}{r|}{\textbf{-0.01 }} & \multicolumn{1}{r}{67.71 } & \multicolumn{1}{r|}{\textbf{0.00 }} & \multicolumn{1}{r}{67.64 } & \multicolumn{1}{r|}{\textbf{-0.07 }} & \multicolumn{1}{r}{67.60 } & \multicolumn{1}{r|}{\textbf{-0.04 }} & \multicolumn{1}{r}{67.60 } & \multicolumn{1}{r|}{\textbf{0.00 }} & \multicolumn{1}{r}{67.60 } & \multicolumn{1}{r|}{\textbf{0.00 }} & \multicolumn{1}{r}{67.60 } & \multicolumn{1}{r|}{\textbf{0.00 }} & \multicolumn{1}{r}{67.60 } & \textbf{0.00 } \\
    \multicolumn{2}{|l|}{WRN-28-10\cite{wang2023better}} & 67.31 & \multicolumn{1}{r|}{67.32 } & \multicolumn{1}{r}{67.30 } & \multicolumn{1}{r|}{\textbf{-0.02 }} & \multicolumn{1}{r}{67.30 } & \multicolumn{1}{r|}{\textbf{0.00 }} & \multicolumn{1}{r}{67.25 } & \multicolumn{1}{r|}{\textbf{-0.05 }} & \multicolumn{1}{r}{67.24 } & \multicolumn{1}{r|}{\textbf{-0.01 }} & \multicolumn{1}{r}{67.24 } & \multicolumn{1}{r|}{\textbf{0.00 }} & \multicolumn{1}{r}{67.24 } & \multicolumn{1}{r|}{\textbf{0.00 }} & \multicolumn{1}{r}{67.24 } & \multicolumn{1}{r|}{\textbf{0.00 }} & \multicolumn{1}{r}{67.24 } & \textbf{0.00 } \\
    \multicolumn{2}{|l|}{WRN-70-16\cite{rebuffi2021fixing}} & 66.59 & \multicolumn{1}{r|}{66.80 } & \multicolumn{1}{r}{66.75 } & \multicolumn{1}{r|}{\textbf{-0.05 }} & \multicolumn{1}{r}{66.72 } & \multicolumn{1}{r|}{\textbf{-0.03 }} & \multicolumn{1}{r}{66.59 } & \multicolumn{1}{r|}{\textbf{-0.13 }} & \multicolumn{1}{r}{66.56 } & \multicolumn{1}{r|}{\textbf{-0.03 }} & \multicolumn{1}{r}{66.55 } & \multicolumn{1}{r|}{\textbf{-0.01 }} & \multicolumn{1}{r}{66.55 } & \multicolumn{1}{r|}{\textbf{0.00 }} & \multicolumn{1}{r}{66.55 } & \multicolumn{1}{r|}{\textbf{0.00 }} & \multicolumn{1}{r}{66.55 } & \textbf{0.00 } \\
    \multicolumn{2}{|l|}{WRN-70-16\cite{gowal2021improving}} & 66.14 & \multicolumn{1}{r|}{66.24 } & \multicolumn{1}{r}{66.19 } & \multicolumn{1}{r|}{\textbf{-0.05 }} & \multicolumn{1}{r}{66.17 } & \multicolumn{1}{r|}{\textbf{-0.02 }} & \multicolumn{1}{r}{66.11 } & \multicolumn{1}{r|}{\textbf{-0.06 }} & \multicolumn{1}{r}{66.09 } & \multicolumn{1}{r|}{\textbf{-0.02 }} & \multicolumn{1}{r}{66.09 } & \multicolumn{1}{r|}{\textbf{0.00 }} & \multicolumn{1}{r}{66.09 } & \multicolumn{1}{r|}{\textbf{0.00 }} & \multicolumn{1}{r}{66.09 } & \multicolumn{1}{r|}{\textbf{0.00 }} & \multicolumn{1}{r}{66.09 } & \textbf{0.00 } \\
    \multicolumn{2}{|l|}{WRN-70-16\cite{gowal2020uncovering}} & 65.89 & \multicolumn{1}{r|}{65.95 } & \multicolumn{1}{r}{65.94 } & \multicolumn{1}{r|}{\textbf{-0.01 }} & \multicolumn{1}{r}{65.92 } & \multicolumn{1}{r|}{\textbf{-0.02 }} & \multicolumn{1}{r}{65.86 } & \multicolumn{1}{r|}{\textbf{-0.06 }} & \multicolumn{1}{r}{65.86 } & \multicolumn{1}{r|}{\textbf{0.00 }} & \multicolumn{1}{r}{65.86 } & \multicolumn{1}{r|}{\textbf{0.00 }} & \multicolumn{1}{r}{65.86 } & \multicolumn{1}{r|}{\textbf{0.00 }} & \multicolumn{1}{r}{65.86 } & \multicolumn{1}{r|}{\textbf{0.00 }} & \multicolumn{1}{r}{65.86 } & \textbf{0.00 } \\
    \multicolumn{2}{|l|}{WRN-A4\cite{huang2023revisiting}} & 65.78 & \multicolumn{1}{r|}{65.87 } & \multicolumn{1}{r}{65.86 } & \multicolumn{1}{r|}{\textbf{-0.01 }} & \multicolumn{1}{r}{65.84 } & \multicolumn{1}{r|}{\textbf{-0.02 }} & \multicolumn{1}{r}{65.70 } & \multicolumn{1}{r|}{\textbf{-0.14 }} & \multicolumn{1}{r}{65.68 } & \multicolumn{1}{r|}{\textbf{-0.02 }} & \multicolumn{1}{r}{65.68 } & \multicolumn{1}{r|}{\textbf{0.00 }} & \multicolumn{1}{r}{65.68 } & \multicolumn{1}{r|}{\textbf{0.00 }} & \multicolumn{1}{r}{65.68 } & \multicolumn{1}{r|}{\textbf{0.00 }} & \multicolumn{1}{r}{65.68 } & \textbf{0.00 } \\
    \multicolumn{2}{|l|}{WRN-106-16\cite{rebuffi2021fixing}} & 64.68 & \multicolumn{1}{r|}{64.69 } & \multicolumn{1}{r}{64.65 } & \multicolumn{1}{r|}{\textbf{-0.04 }} & \multicolumn{1}{r}{64.64 } & \multicolumn{1}{r|}{\textbf{-0.01 }} & \multicolumn{1}{r}{64.57 } & \multicolumn{1}{r|}{\textbf{-0.07 }} & \multicolumn{1}{r}{64.56 } & \multicolumn{1}{r|}{\textbf{-0.01 }} & \multicolumn{1}{r}{64.56 } & \multicolumn{1}{r|}{\textbf{0.00 }} & \multicolumn{1}{r}{64.55 } & \multicolumn{1}{r|}{\textbf{-0.01 }} & \multicolumn{1}{r}{64.55 } & \multicolumn{1}{r|}{\textbf{0.00 }} & \multicolumn{1}{r}{64.55 } & \textbf{0.00 } \\
    \midrule
    \multicolumn{20}{|c|}{} \\
    \midrule
    \multicolumn{2}{|c|}{\textbf{Defense}} & \multicolumn{1}{c|}{\textbf{AA}} & \multicolumn{2}{c|}{\textbf{+$\text{PGD}_\text{alt}$}} & \multicolumn{2}{c|}{\textbf{+$\text{APGD}_\text{ce}$}} & \multicolumn{2}{c|}{\textbf{+$\text{APGD}_\text{dlr}$}} & \multicolumn{2}{c|}{\textbf{+$\text{APGD}_\text{mg}$}} & \multicolumn{2}{c|}{\textbf{+$\text{APGD}_\text{mi}$}} & \multicolumn{2}{c|}{\textbf{+$\text{APGDT}_\text{dlr}$}} & \multicolumn{2}{c|}{\textbf{+$\text{APGDT}_\text{mg}$}} & \multicolumn{2}{c|}{\textbf{+FAB}} & \multicolumn{1}{l|}{\textbf{diff}} \\
    \midrule
    \multicolumn{2}{|l|}{RWRN-70-16\cite{peng2023robust}} & 71.10 & 71.00  & \textbf{0.00 } & 71.00  & \textbf{0.00 } & 71.00  & \textbf{0.00 } & 71.00  & \textbf{0.00 } & 71.00  & \textbf{0.00 } & 71.00  & \textbf{0.00 } & 71.00  & \textbf{0.00 } & 71.00  & \textbf{0.00 } & \textbf{-0.10 } \\
    \multicolumn{2}{|l|}{WRN-70-16\cite{wang2023better}} & 70.70 & 70.57  & \textbf{0.00 } & 70.57  & \textbf{0.00 } & 70.57  & \textbf{0.00 } & 70.57  & \textbf{0.00 } & 70.57  & \textbf{0.00 } & 70.56  & \textbf{-0.01 } & 70.56  & \textbf{0.00 } & 70.56  & \textbf{0.00 } & \textbf{-0.14 } \\
    \multicolumn{2}{|l|}{Mixing\cite{bai2023improving}} & 68.06 & 67.23  & \textbf{-0.14 } & 67.23  & \textbf{0.00 } & 67.19  & \textbf{-0.04 } & 67.16  & \textbf{-0.03 } & 67.15  & \textbf{-0.01 } & 67.07  & \textbf{-0.08 } & 67.07  & \textbf{0.00 } & 67.07  & \textbf{0.00 } & \textbf{-0.99 } \\
    \multicolumn{2}{|l|}{WRN-28-10\cite{cui2023decoupled}} & 67.75 & 67.60  & \textbf{0.00 } & 67.60  & \textbf{0.00 } & 67.59  & \textbf{-0.01 } & 67.59  & \textbf{0.00 } & 67.59  & \textbf{0.00 } & 67.58  & \textbf{-0.01 } & 67.57  & \textbf{-0.01 } & 67.57  & \textbf{0.00 } & \textbf{-0.18 } \\
    \multicolumn{2}{|l|}{WRN-28-10\cite{wang2023better}} & 67.31 & 67.24  & \textbf{0.00 } & 67.24  & \textbf{0.00 } & 67.24  & \textbf{0.00 } & 67.24  & \textbf{0.00 } & 67.24  & \textbf{0.00 } & 67.20  & \textbf{-0.04 } & 67.20  & \textbf{0.00 } & 67.20  & \textbf{0.00 } & \textbf{-0.11 } \\
    \multicolumn{2}{|l|}{WRN-70-16\cite{rebuffi2021fixing}} & 66.59 & 66.55  & \textbf{0.00 } & 66.55  & \textbf{0.00 } & 66.55  & \textbf{0.00 } & 66.55  & \textbf{0.00 } & 66.55  & \textbf{0.00 } & 66.52  & \textbf{-0.03 } & 66.52  & \textbf{0.00 } & 66.52  & \textbf{0.00 } & \textbf{-0.07 } \\
    \multicolumn{2}{|l|}{WRN-70-16\cite{gowal2021improving}} & 66.14 & 66.09  & \textbf{0.00 } & 66.08  & \textbf{-0.01 } & 66.07  & \textbf{-0.01 } & 66.07  & \textbf{0.00 } & 66.07  & \textbf{0.00 } & 66.04  & \textbf{-0.03 } & 66.04  & \textbf{0.00 } & 66.04  & \textbf{0.00 } & \textbf{-0.10 } \\
    \multicolumn{2}{|l|}{WRN-70-16\cite{gowal2020uncovering}} & 65.89 & 65.86  & \textbf{0.00 } & 65.86  & \textbf{0.00 } & 65.83  & \textbf{-0.03 } & 65.82  & \textbf{-0.01 } & 65.82  & \textbf{0.00 } & 65.82  & \textbf{0.00 } & 65.82  & \textbf{0.00 } & 65.82  & \textbf{0.00 } & \textbf{-0.07 } \\
    \multicolumn{2}{|l|}{WRN-A4\cite{huang2023revisiting}} & 65.78 & 65.68  & \textbf{0.00 } & 65.68  & \textbf{0.00 } & 65.68  & \textbf{0.00 } & 65.68  & \textbf{0.00 } & 65.68  & \textbf{0.00 } & 65.68  & \textbf{0.00 } & 65.68  & \textbf{0.00 } & 65.68  & \textbf{0.00 } & \textbf{-0.10 } \\
    \multicolumn{2}{|l|}{WRN-106-16\cite{rebuffi2021fixing}} & 64.68 & 64.55  & \textbf{0.00 } & 64.55  & \textbf{0.00 } & 64.53  & \textbf{-0.02 } & 64.52  & \textbf{-0.01 } & 64.52  & \textbf{0.00 } & 64.52  & \textbf{0.00 } & 64.52  & \textbf{0.00 } & 64.52  & \textbf{0.00 } & \textbf{-0.16 } \\
    \bottomrule
    \end{tabular}%
    \end{adjustbox}
    \caption{ 
  The CIFAR-10 models' robustness (\%) evaluated by a cascade ensemble attack that starts with our PMA and then includes more attacks (to the right) sequentially until the bottom right of the second row. Within each column, the left and right numbers show the absolute value and robustness decreases, respectively. The final "diff" column marks the overall robustness difference between AA and the full ensemble of attacks (our PMA + 16 other attacks).
  }
  \label{tab:PMA+n}%
  \vspace{-5pt}
\end{table*}%

\vspace{-10pt}

\begin{table*}[h]
  \centering
  \begin{adjustbox}{width=0.9\linewidth}
  \setlength{\extrarowheight}{-2pt}
    \begin{tabular}{rrrrrrrrrrrrll|c|ccc}
    \toprule
    \multicolumn{2}{c|}{\textbf{Defense}} & \multicolumn{1}{c|}{\textbf{Clean}} & \multicolumn{1}{c}{\textbf{AA}} & \multicolumn{1}{c}{\textbf{PMA+}} & \multicolumn{1}{c|}{\textbf{diff}} & \multicolumn{2}{c|}{\textbf{Defense}} & \multicolumn{1}{c|}{\textbf{Clean}} & \multicolumn{1}{c}{\textbf{AA}} & \multicolumn{1}{c}{\textbf{PMA+}} & \multicolumn{1}{c|}{\textbf{diff}} & \multicolumn{2}{c|}{\textbf{Defense}} & \textbf{Clean} & \textbf{AA} & \textbf{PMA+} & \textbf{diff} \\
    \midrule
    \multicolumn{6}{c|}{\textbf{CIFAR10}}         & \multicolumn{6}{c|}{\textbf{CIFAR100}}        & \multicolumn{6}{c}{\textbf{ImageNet-1k}} \\
    \midrule
    \multicolumn{2}{l|}{RWRN-70-16\cite{peng2023robust}} & \multicolumn{1}{c|}{93.27} & \multicolumn{1}{c}{71.10} & \multicolumn{1}{c}{\textbf{71.05}} & \multicolumn{1}{c}{\textbf{-0.05}} & \multicolumn{2}{l|}{WRN-70-16\cite{wang2023better}} & \multicolumn{1}{c|}{75.23} & \multicolumn{1}{c}{42.68} & \multicolumn{1}{c}{\textbf{42.63}} & \multicolumn{1}{c}{\textbf{-0.05}} & \multicolumn{2}{l|}{Swin-L\cite{liu2023comprehensive}} & 78.18 & 57.26 & \textbf{57.20} & \textbf{-0.06} \\
    \multicolumn{2}{l|}{WRN-70-16\cite{wang2023better}} & \multicolumn{1}{c|}{93.25} & \multicolumn{1}{c}{70.70} & \multicolumn{1}{c}{\textbf{70.61}} & \multicolumn{1}{c}{\textbf{-0.09}} & \multicolumn{2}{l|}{WRN-28-10\cite{cui2023decoupled}} & \multicolumn{1}{c|}{73.83} & \multicolumn{1}{c}{39.20} & \multicolumn{1}{c}{\textbf{39.17}} & \multicolumn{1}{c}{\textbf{-0.03}} & \multicolumn{2}{l|}{Mixing\cite{bai2024mixednuts}} & 81.10 & 58.31 & \textbf{58.22} & \textbf{-0.09} \\
    \multicolumn{2}{l|}{Mixing\cite{bai2023improving}} & \multicolumn{1}{c|}{95.23} & \multicolumn{1}{c}{68.06} & \multicolumn{1}{c}{\textbf{67.91}} & \multicolumn{1}{c}{\textbf{-0.15}} & \multicolumn{2}{l|}{WRN-28-10\cite{wang2023better}} & \multicolumn{1}{c|}{72.58} & \multicolumn{1}{c}{38.79} & \multicolumn{1}{c}{\textbf{38.77}} & \multicolumn{1}{c}{\textbf{-0.02}} & \multicolumn{2}{l|}{ConvNeXt-L\cite{liu2023comprehensive}} & 77.48 & 56.42 & \textbf{56.36} & \textbf{-0.06} \\
    \multicolumn{2}{l|}{WRN-28-10\cite{cui2023decoupled}} & \multicolumn{1}{c|}{92.16} & \multicolumn{1}{c}{67.75} & \multicolumn{1}{c}{\textbf{67.63}} & \multicolumn{1}{c}{\textbf{-0.12}} & \multicolumn{2}{l|}{WRN-70-16\cite{gowal2020uncovering}} & \multicolumn{1}{c|}{69.15} & \multicolumn{1}{c}{36.89} & \multicolumn{1}{c}{\textbf{36.85}} & \multicolumn{1}{c}{\textbf{-0.04}} & \multicolumn{2}{l|}{ConvNeXt-L+CS\cite{singh2024revisiting}} & 76.79 & 55.86 & \textbf{55.81} & \textbf{-0.05} \\
    \multicolumn{2}{l|}{WRN-28-10\cite{wang2023better}} & \multicolumn{1}{c|}{92.44} & \multicolumn{1}{c}{67.31} & \multicolumn{1}{c}{\textbf{67.20}} & \multicolumn{1}{c}{\textbf{-0.11}} & \multicolumn{2}{l|}{XCiT-L12\cite{debenedetti2023light}} & \multicolumn{1}{c|}{70.77} & \multicolumn{1}{c}{35.04} & \multicolumn{1}{c}{\textbf{34.95}} & \multicolumn{1}{c}{\textbf{-0.09}} & \multicolumn{2}{l|}{Swin-B\cite{liu2023comprehensive}} & 76.22 & 54.30 & \textbf{54.24} & \textbf{-0.06} \\
    \multicolumn{2}{l|}{WRN-70-16\cite{rebuffi2021fixing}} & \multicolumn{1}{c|}{92.23} & \multicolumn{1}{c}{66.59} & \multicolumn{1}{c}{\textbf{66.57}} & \multicolumn{1}{c}{\textbf{-0.02}} & \multicolumn{2}{l|}{WRN-70-16\cite{rebuffi2021fixing}} & \multicolumn{1}{c|}{63.56} & \multicolumn{1}{c}{34.68} & \multicolumn{1}{c}{\textbf{34.63}} & \multicolumn{1}{c}{\textbf{-0.05}} & \multicolumn{2}{l|}{ConvNeXt-B+CS\cite{singh2024revisiting}} & 75.46 & 53.84 & \textbf{53.80} & \textbf{-0.04} \\
    \multicolumn{2}{l|}{WRN-70-16\cite{gowal2021improving}} & \multicolumn{1}{c|}{88.74} & \multicolumn{1}{c}{66.14} & \multicolumn{1}{c}{\textbf{66.09}} & \multicolumn{1}{c}{\textbf{-0.05}} & \multicolumn{2}{l|}{XCiT-M12\cite{debenedetti2023light}} & \multicolumn{1}{c|}{69.20} & \multicolumn{1}{c}{34.20} & \multicolumn{1}{c}{\textbf{34.13}} & \multicolumn{1}{c}{\textbf{-0.07}} & \multicolumn{2}{l|}{ConvNeXt-B\cite{liu2023comprehensive}} & 76.38 & 54.04 & \textbf{53.99} & \textbf{-0.05} \\
    \multicolumn{2}{l|}{WRN-70-16\cite{gowal2020uncovering}} & \multicolumn{1}{c|}{91.10} & \multicolumn{1}{c}{65.89} & \multicolumn{1}{c}{\textbf{65.86}} & \multicolumn{1}{c}{\textbf{-0.03}} & \multicolumn{2}{l|}{WRN-70-16\cite{pang2022robustness}} & \multicolumn{1}{c|}{65.56} & \multicolumn{1}{c}{33.04} & \multicolumn{1}{c}{\textbf{33.00}} & \multicolumn{1}{c}{\textbf{-0.04}} & \multicolumn{2}{l|}{ViT-B+CS\cite{singh2024revisiting}} & 76.12 & 52.66 & \textbf{52.57} & \textbf{-0.09} \\
\cmidrule{7-12}    \multicolumn{2}{l|}{WRN-A4\cite{huang2023revisiting}} & \multicolumn{1}{c|}{91.59} & \multicolumn{1}{c}{65.78} & \multicolumn{1}{c}{\textbf{65.72}} & \multicolumn{1}{c}{\textbf{-0.06}} &       &       &       &       &       &       & \multicolumn{2}{l|}{ConvNeXt-S+CS\cite{singh2024revisiting}} & 73.37 & 49.65 & \textbf{49.58} & \textbf{-0.07} \\
    \multicolumn{2}{l|}{WRN-106-16\cite{rebuffi2021fixing}} & \multicolumn{1}{c|}{88.50} & \multicolumn{1}{c}{64.68} & \multicolumn{1}{c}{\textbf{64.60}} & \multicolumn{1}{c}{\textbf{-0.08}} &       &       &       &       &       &       & \multicolumn{2}{l|}{ConvNeXt-T+CS\cite{singh2024revisiting}} & 72.45 & 47.60 & \textbf{47.56} & \textbf{-0.04} \\
\cmidrule{1-6}          &       &       &       &       &       &       &       &       &       &       &       & \multicolumn{2}{l|}{RWRN-101-2\cite{peng2023robust}} & 73.45 & 48.96 & \textbf{48.93} & \textbf{-0.03} \\
    \bottomrule
    \end{tabular}%
  \end{adjustbox}
  \caption{A robustness (\%) comparison between PMA+APGDT (PMA+) and the AA across CIFAR-10,  CIFAR-100, and ImageNet-1k datasets. The "diff" column reports the robustness decrease by PMA+ compared to AA. The best results are boldfaced.}
  \label{tab:PMA+}%
\end{table*}%

The results of the \textbf{\emph{cascade ensemble}} are reported in Table \ref{tab:PMA+n}. As can be seen, when more attacks are appended to the ensemble, it keeps reducing the models' robustness to a lower level than that measured by AA, leading to more and more accurate robustness evaluations. The final "diff" column indicates the best result one can obtain with current attacks. 
The maximum robustness difference compared to AA was observed for the top-3 defense model \cite{bai2023improving} which is  0.99\%. Arguably, there exists a trade-off between effectiveness and efficiency. Our results indicate that the robustness difference is within 1\% for the top-ranked CIFRA-10 defense models when applying an ensemble of 17 attacks vs. 1 single attack to evaluate the robustness. As such, we recommend using the \textbf{\emph{PMA+1 ensemble}}, PMA+APGDT (denoted as \textbf{PMA+}) to be more specific, to evaluate future defenses. A performance comparison between \textbf{PMA+} and AA can be found in Table \ref{tab:PMA+}. It is clear that \textbf{PMA+} outperforms AA in all cases, but \textbf{PMA+} only takes ~25\% of the running time of AA.

\subsection{Million-Scale Robustness Evaluation}
Here, we aim to scale adversarial robustness evaluation up to the million-scale for adversarially trained models on ImageNet. To this end, we first construct a large-scale evaluation dataset named CC1M with 1 million images selected from the CC3M \cite{sharma2018conceptual} dataset, which is an (image, caption) paired dataset comprising diverse objects, scenes, and visual concepts.  We first remove the "unavailable" images, i.e., images showing `this image is unavailable', due to a large number of expired image URLs. Next, we remove the noisy images that do not contain any semantic content, such as random icons. Finally, we apply the Local Intrinsic Dimensionality (LID) \cite{houle2017local1} to remove anomaly images. The LID is a representation-space metric that has been shown to be able to detect adversarial images \cite{ma2018characterizing}, backdoor images \cite{dolatabadi2022collider,huang2025detecting}, or low-quality images that are detrimental to self-supervised contrastive learning \cite{huang2024ldreg}. Images with deviated LID scores from the average are often viewed as outliers that are geometrically far away from normal data manifold. Therefore, we adopt the Median Absolute Deviation (MAD) with LID as the final score to filter out the outlier (unusual) images. 
Specifically, we first compute the LID scores for images in CC3M based on their CLIP embeddings, calculate the median of the LID scores, and then select 1 million images with LID scores close to the median. 


\begin{figure}[h]
    \centering
    \includegraphics[width=0.45\textwidth]{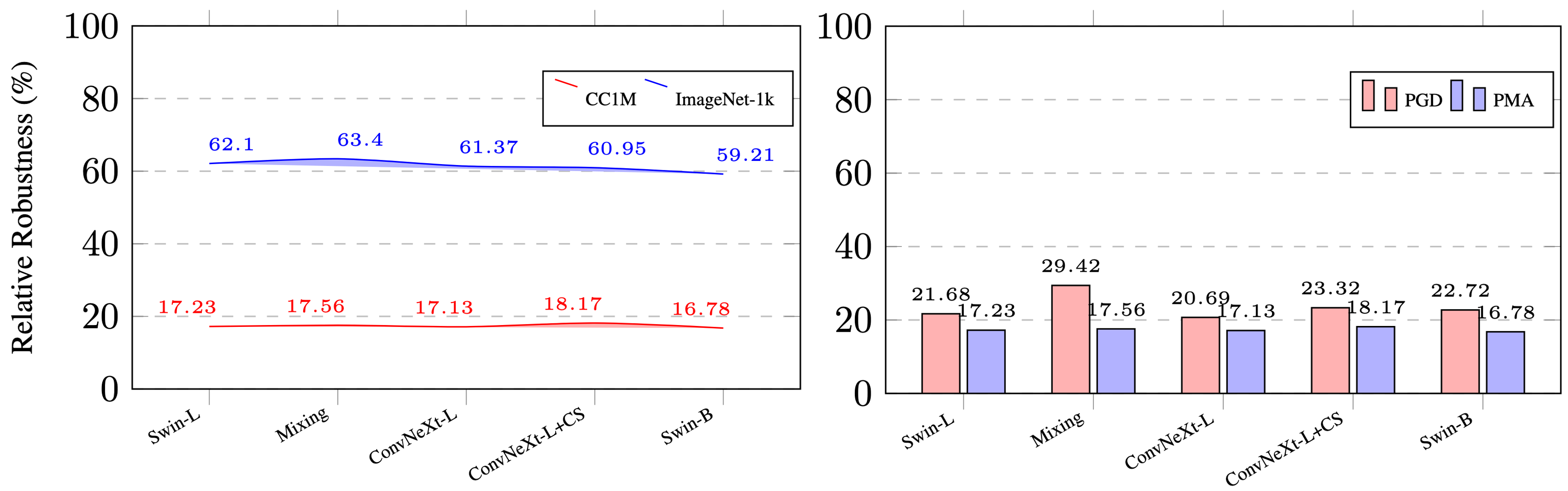}
    \caption{Relative robustness evaluation on CC1M and ImageNet-1k test set.  $x$-axis: top-5 (on RobustBench leaderboard) defense models adversarially trained on ImageNet; $y$-axis: relative robustness. \textbf{Left}: The red and blue lines represent the \emph{relative robustness} measured on CC1M and ImageNet-1k test set, respectively. \textbf{Right}:The red and blue bars represent the \emph{relative robustness} measured on CC1M by $\text{PGD}_{\text{ce}}$ and our PMA, respectively.}
    \label{fig:robustness_attack}
\end{figure}


To verify the robustness of a model on a large-scale unlabeled dataset, we define a new robustness metric called \textbf{Relative Robustness}, which replaces the ground truth label in the standard robustness by the predicted labels on the clean images. The attack is considered successful when the model's prediction on the adversarial image does not match its prediction on the clean image. Note that \emph{relative robustness} is a weaker robustness measurement than the standard robustness as it also considers the incorrect predictions. We evaluate the relative robustness of the top 5 models on the ImageNet leaderboard from RobustBench using 100 steps PMA. The evaluation was conducted on both the CC1M and ImageNet-1k datasets and the results are shown in Figure \ref{fig:robustness_attack}. The relative robustness of these models decreases drastically from above 59\% to below 19\%. And the relative rankings between the models have also changed. This suggests a considerable robustness gap between small-scale evaluation and large-scale evaluation, revealing the vulnerabilities of ImageNet pre-trained robust models in broader applications.

In Figure \ref{fig:robustness_attack}, we compare the relative robustness measured on CC1M using our PMA and the $\text{PGD}_{\text{ce}}$ method, both with 100 steps. For a comprehensive comparison with other methods, please see Appendix F.
It should be noted that, due to the high computational cost of AA, we did not include a comparison with AA in this experiment. As can be observed, the robustness evaluated by PMA is approximately 3.56\% to 11.86\% lower than that evaluated by $\text{PGD}_{\text{ce}}$. This verifies the advantage and reliability of our method in large-scale robustness evaluation.

\label{sec:con}

\section{Limitation}
While PMA outperforms all individual attacks and the PMA+ ensemble achieves state-of-the-art effectiveness, it still has some limitations. Although PMA+ is significantly more efficient than AutoAttack (AA), it still requires considerable time for large-scale evaluations. Additionally, our approach is currently limited to white-box settings, posing challenges for extending its applicability to black-box attack scenarios.




\section{Conclusion}


In this paper, we studied the problem of white-box robustness evaluation and introduced a novel individual attack, Probability Margin Attack (PMA), which is guided by the newly proposed Probability Margin (PM) loss. We analyzed the relationship between PM loss and several widely used loss functions, including cross-entropy loss (both targeted and untargeted), difference of logits ratio (DLR), and margin loss. Through empirical evaluation, we demonstrated the superior performance of the PM loss and the PMA attack, highlighting their effectiveness. Additionally, we explored the potential for developing an ensemble attack that outperforms AutoAttack (AA) in both effectiveness and speed. Finally, we conducted a million-scale white-box adversarial robustness evaluation on adversarially trained ImageNet models, revealing a significant robustness gap between small-scale and large-scale evaluations.

\vspace{-10pt}

\paragraph{Acknowledgements} This work is in part supported by the National Key R\&D Program of China (Grant No. 2022ZD0160103) and the National Natural Science Foundation of China (Grant No. 62276067).
The computations in this research were performed using the CFFF platform of Fudan University.


\twocolumn
{
    \small
    \bibliographystyle{IEEEtran}  
    \bibliography{main}
}

\clearpage
\setcounter{page}{1}
\maketitlesupplementary


\section{Introduction}
Due to the page limitation of the paper, we further illustrate our method in this supplementary material, which includes the following sections: 
1)Visual illustrations of the attacked images.
2) A detailed analysis of the quantitative results for hyperparameters $K'$ and $n$; 
3) A comparison of experimental results between the PMA method and the AAA and ACG methods; 
4) A comparison of experimental results between the traditional SGD+sign update strategy and optimizer-based strategies; 
5) A detailed examination of the ablation results for $P_{max}$ and $P_y$ weights; 
6) Detailed results of the million-scale adversarial robustness evaluation between the PMA method and other methods.
7) Supplementary experiments on CLIP.

\section{isual illustrations of the attacked images.}

\begin{figure}[htbp]
    \vspace{-4pt}
    \centering
    \includegraphics[width=\linewidth]{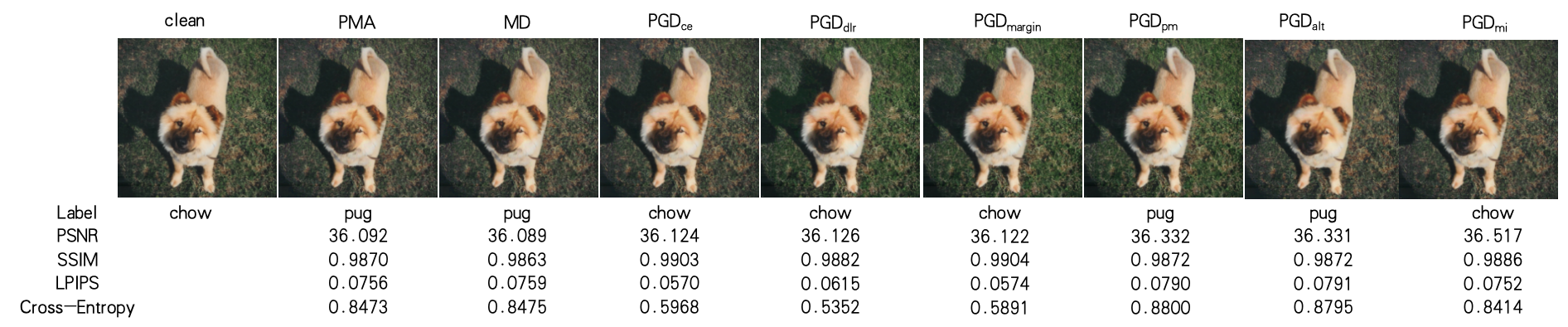} 
    \caption{Visual illustrations of the attacked images.}
    \label{fig:example}
\end{figure}

Fig.~\ref{fig:example} visualizes adversarial examples generated by different attacks, along with their predicted labels. We evaluate their quality using PSNR, SSIM, and LPIPS, and report cross-entropy loss between adversarial predictions and ground-truth labels, demonstrating both visual and functional impact.

\section{Detailed quantitative results of hyperparameter $K'$ and $n$} 

\begin{table}[htbp]
\centering
\caption{The models’ robustness (\%) evaluated on different $K'$ values. The best results are boldfaced.}
\begin{adjustbox}{width=1\linewidth}
\begin{tabular}{l|c|r|r|r|r|r}
\hline
\textbf{Dataset} & \textbf{Model} & $\boldsymbol{K' = 15}$ & $\boldsymbol{K' = 20}$ & $\boldsymbol{K' = 25}$ & $\boldsymbol{K' = 30}$ & $\boldsymbol{K' = 35}$ \\
\hline
CIFAR10 & WRN-28-10\cite{cui2023decoupled} & 67.79 & 67.76 & \textbf{67.72} & 67.77 & 67.77 \\
CIFAR10 & WRN-28-10\cite{wang2023better} & 67.35 & 67.34 & \textbf{67.33} & 67.36 & 67.35 \\
CIFAR10 & RWRN-70-16\cite{peng2023robust} & 71.13 & 71.14 & \textbf{71.1} & 71.14 & 71.16 \\
ImageNet & ViT-B+CS\cite{singh2024revisiting} & 52.84 & 52.83 & \textbf{52.82} & 52.86 & 54.43 \\
ImageNet & Swin-B\cite{liu2023comprehensive} & 54.41 & 54.41 & \textbf{54.41} & 54.41 & 54.43 \\
ImageNet & ConvNeXt-S+CS\cite{singh2024revisiting} & 49.74 & 49.74 & \textbf{49.74} & 49.75 & 49.8 \\
\hline
\end{tabular}
\end{adjustbox}
\label{tab:k_results}
\end{table}


\begin{table}[htbp]
\centering
\caption{The models’ robustness (\%) evaluated on different $\text{n}$ values. The best results are boldfaced.}
\begin{adjustbox}{width=0.7\linewidth}
\begin{tabular}{l|c|r|r|r}
\hline
\textbf{Dataset} & \textbf{Model} & $\boldsymbol{n = 1}$ & $\boldsymbol{n = 2}$ & $\boldsymbol{n = 5}$ \\
\hline
CIFAR10 & WRN-28-10\cite{cui2023decoupled} & \textbf{67.72} & 67.8 & 68.18 \\
CIFAR10 & WRN-28-10\cite{wang2023better} & \textbf{67.33} & 67.35 & 67.65 \\
CIFAR10 & RWRN-70-16\cite{peng2023robust} & \textbf{71.1} & 71.14 & 71.48 \\
ImageNet & ViT-B+CS\cite{singh2024revisiting} & \textbf{52.82} & 52.9 & 53.17 \\
ImageNet & Swin-B\cite{liu2023comprehensive} & \textbf{54.41} & 54.45 & 54.68 \\
ImageNet & ConvNeXt-S+CS\cite{singh2024revisiting} &\textbf{49.74} & 49.75 & 49.97 \\
\hline
\end{tabular}
\end{adjustbox}
\label{tab:n_results}
\end{table}


In our study, we conducted a quantitative analysis of the hyperparameters $K'$ and $n$ for the PMA method. We tested three models from the CIFAR10 dataset and three defense models from ImageNet, with a total of 100 iterations set for the tests. The results are presented by evaluating the robust accuracy of different defense models under various attacks.

For the quantitative investigation of $K'$, we compared five different values: 15, 20, 25, 30, and 35. The results, as shown in Table \ref{tab:k_results}, indicate that the optimal performance is achieved when $K$ is set to 25.

In the quantitative study of $n$, we set three different numbers of restarts: 1, 2, and 5, ensuring a total of 100 iterations. The results, as shown in Table \ref{tab:n_results}, suggest that the best performance is obtained when $n$ is set to 1.

\section{Comparison of experimental results between PMA method and AAA, ACG methods}

In this comparative analysis, we evaluated the AAA and ACG methods alongside our PMA method. Seven defense models from the CIFAR10 dataset were subjected to a constraint of 100 attack steps. The outcomes are detailed by assessing the robust accuracy of these models when confronted with diverse attack scenarios. As depicted in Table \ref{tab:AAA&ACG_results}, our approach consistently outperformed the others, demonstrating superior effectiveness.


\begin{table}[htbp]
\centering
\caption{The robustness (\%) of different models on the CIFAR10 dataset, as evaluated by PMA, AAA, and ACG attacks.}
\begin{adjustbox}{width=0.8\linewidth}
\begin{tabular}{l|c|r|r|r}
\hline
\textbf{Dataset} & \textbf{Model} & \textbf{PMA} & \textbf{AAA} & \textbf{ACG} \\
\hline
CIFAR10 & WRN-28-10\cite{cui2023decoupled} & \textbf{67.72} & 68.85 & 68.63 \\
CIFAR10 & WRN-28-10\cite{wang2023better} & \textbf{67.33} & 68.49 & 68.26 \\
CIFAR10 & WRN-70-16\cite{gowal2020uncovering} & \textbf{65.95} & 71.27 & 69.39 \\
CIFAR10 & Mixing\cite{bai2023improving} & \textbf{68.43} & 71.27 & 69.39 \\
CIFAR10 & WRN-70-16\cite{rebuffi2021fixing} & \textbf{66.80} & 68.18 & 67.78 \\
CIFAR10 & WRN-106-16\cite{rebuffi2021fixing} & \textbf{64.69} & 65.84 & 65.62 \\
CIFAR10 & WRN-70-16\cite{wang2023better} & \textbf{70.67} & 71.18 & 71.58 \\
\hline
\end{tabular}
\end{adjustbox}
\label{tab:AAA&ACG_results}
\end{table}

\begin{table*}[h]
\centering
\caption{The model's robustness(\%) evaluated by individual attacks. The best results are boldfaced.}
\begin{adjustbox}{width=0.9\linewidth}
\begin{tabular}{l|c|r|r|r|r|r|r|r|r}
\hline
\textbf{Dataset} & \textbf{Model} & \textbf{$\textbf{PGD}_\text{ce}$} & \textbf{$\textbf{PGD}_\text{dlr}$} & \textbf{$\textbf{PGD}_\text{mg}$} & \textbf{$\textbf{PGD}_\text{pm}$} & \textbf{$\textbf{PGD}_\text{alt}$} & \textbf{$\textbf{PGD}_\text{mi}$} & \textbf{$\text{MD}$} & \textbf{$\text{PMA}$} \\
\hline
CIFAR10 & WRN-28-10\cite{cui2023decoupled} & 72.26/\textbf{70.65} & 69.88/\textbf{68.65} & 69.65/\textbf{68.62} & 69.52/\textbf{68.47} & 68.83/\textbf{67.88} & 69.1/\textbf{67.95} & 71.61/\textbf{67.79} & 71.9/\textbf{67.72} \\
CIFAR10 & WRN-28-10\cite{wang2023better} & 71.72/\textbf{70.31} & 69.55/\textbf{68.31} & 69.22/\textbf{68.22} & 69.19/\textbf{68.10} & 68.42/\textbf{67.46} & 68.76/\textbf{67.51} & 71.46/\textbf{67.42} & 71.46/\textbf{67.33} \\
CIFAR10 & RWRN-70-16\cite{peng2023robust} & 75.14/\textbf{73.98} & 73.43/\textbf{72.03} & 73.09/\textbf{71.94} & 73.03/\textbf{71.16} & 72.25/\textbf{71.15} & 72.49/\textbf{71.25} & 74.9/\textbf{71.14} & 74.9/\textbf{71.10} \\
ImageNet & ViT-B+CS\cite{singh2024revisiting} & 57.39/\textbf{55.34} & 56.84/\textbf{55.52} & 55.93/\textbf{55.01} & 54.72/\textbf{53.61} & 54.21/\textbf{53.00} & 55.7/\textbf{53.19} & 55.48/\textbf{53.34} & 55.10/\textbf{52.82} \\
ImageNet & Swin-B\cite{liu2023comprehensive} & 58.5/\textbf{57.26} & 57.57/\textbf{56.69} & 56.92/\textbf{56.28} & 55.98/\textbf{54.93} & 55.27/\textbf{54.55} & 56.36/\textbf{54.57} & 59.31/\textbf{54.48} & 61.04/\textbf{54.57} \\
ImageNet & ConvNeXt-S+CS\cite{singh2024revisiting} & 53.58/\textbf{52.69} & 53.63/\textbf{52.72} & 52.63/\textbf{51.94} & 51.5/\textbf{50.38} & 50.79/\textbf{49.93} & 51.98/\textbf{49.98} & 53.71/\textbf{50.12} & 52.97/\textbf{49.74} \\
\hline
\end{tabular}
\end{adjustbox}
\label{tab:adam}
\end{table*}

\begin{table*}[htpb]
\centering
\caption{The models's robustness results across various methods on CC1M, with the best performances in bold.}
\begin{adjustbox}{width=0.8\linewidth}
\begin{tabular}{l|c|r|r|r|r|r|r|r|r|r}
\hline
\textbf{Dataset} & \textbf{Model} & \textbf{$\textbf{PGD}_\text{ce}$} & \textbf{$\textbf{PGD}_\text{dlr}$} & \textbf{$\textbf{PGD}_\text{mg}$} & \textbf{$\textbf{PGD}_\text{pm}$} & \textbf{$\textbf{PGD}_\text{alt}$} & \textbf{$\textbf{PGD}_\text{mi}$} & \textbf{$\text{MD}$} & \textbf{$\text{PMA}$} & \textbf{$\text{AA}$} \\
\hline
ImageNet & Swin-L\cite{liu2023comprehensive} & 20.66 & 19.82 & 19.06 & 17.38 & 16.68 & 16.72 & 16.68 & \textbf{16.54} & \textbf{16.3} \\
ImageNet & Mixing\cite{bai2024mixednuts} & 29.46 & 20.9 & 19.06 & 18.32 & 18.42 & 21.48 & 17.48 & \textbf{17.4} & \textbf{16.64} \\
ImageNet & ConvNeXt-L\cite{liu2023comprehensive} & 20.26 & 20.9 & 20.3 & 18.5 & 17.4 & 17.3 & 17.3 & \textbf{16.92} & \textbf{16.78} \\
ImageNet & ConvNeXt-L+CS\cite{singh2024revisiting} & 22.76 & 21.09 & 21 & 18.88 & 18.2 & 18.14 & 19.02 & \textbf{18} & \textbf{17.58} \\
ImageNet & Swin-B\cite{liu2023comprehensive} & 22.24 & 19.2 & 16.9 & 16.78 & 16.98 & 16.8 & 16.9 & \textbf{16.64} & \textbf{16.54} \\
\hline
\end{tabular}
\end{adjustbox}
\label{tab:model_robustness_results}
\end{table*}

\section{Comparison of experimental results between traditional SGD+sign update strategy and optimizer-based strategies}

In our preliminary experiments, we adopted the SGD+sign update strategy, forgoing the integration of an optimizer. To extend our analysis, this section introduces comparative experiments with optimizer-based approaches, focusing on the widely recognized Adam optimizer.

We evaluated three models from the CIFAR10 dataset and three defense models from ImageNet, each subjected to a total of 100 iterations. For the optimizer configuration, we employed the tanh function to scale the noise within the interval [-1, 1], multiplied this scaled noise by the perturbation magnitude, and subsequently added it to the original image. Clipping was applied to ensure pixel values remained within the permissible range. The initial learning rate was set to 0.05, with $\beta_1 = 0.9$ and $\beta_2 = 0.99$.

The ensuing robustness outcomes, presented as a comparison between 'Adam' and 'SGD+sign' in Table \ref{tab:adam}, indicate that the Adam optimizer does not significantly enhance the efficacy of the attack. However, in this context, our pm loss, as implemented in the PMA method, consistently demonstrated superior performance. This underscores the critical importance of identifying more reliable optimization directions in the domain of adversarial attacks.

\section{Detailed ablation results of $P_{max}$ and $P_y$ weights}

\begin{table}[htbp]
\centering
\caption{The robustness (\%) of the models, evaluated using the \textbf{$\text{PGD}_\text{pm}$} attack with varying $\beta$ values, on the CIFAR10 and ImageNet datasets.}
\begin{adjustbox}{width=1\linewidth}
\begin{tabular}{l|c|r|r|r|r|r}
\hline
\textbf{Dataset} & \textbf{Model} & $\boldsymbol{\beta = 0.5}$ & $\boldsymbol{\beta = 0.75}$ & $\boldsymbol{\beta = 1}$ & $\boldsymbol{\beta = 1.25}$ & $\boldsymbol{\beta = 1.5}$ \\
\hline
CIFAR10 & WRN-28-10\cite{cui2023decoupled} & 68.66 & \textbf{68.47} & 68.47 & 68.48 & 68.54 \\
CIFAR10 & WRN-28-10\cite{wang2023better} & 68.11 & \textbf{68.09} & 68.1 & 68.16 & 68.26 \\
CIFAR10 & RWRN-70-16\cite{peng2023robust} & 71.95 & \textbf{71.74} & 71.76 & 71.79 & 71.87 \\
ImageNet & ViT-B+CS\cite{singh2024revisiting} & 53.48 & \textbf{53.48} & 53.61 & 53.8 & 54.04 \\
ImageNet & Swin-B\cite{liu2023comprehensive} & 54.94 & \textbf{54.83} & 54.93 & 55.11 & 55.31 \\
ImageNet & ConvNeXt-S+CS\cite{singh2024revisiting} & 50.38 & \textbf{50.26} & 50.38 & 50.56 & 50.82 \\
\hline
\end{tabular}
\end{adjustbox}
\label{tab:pgd_pm_results}
\end{table}


In this section, we present a quantitative analysis of the weights associated with $P_{max}$ and $P_y$ within the PMA and $PGD_{pm}$ approaches. We utilized a weighted formulation of the PM loss, defined as $L_{pm} = \beta \cdot P_{max} - P_y$, to evaluate both the PGD and PMA methods. The evaluation encompassed three models from the CIFAR10 dataset and three defense models from ImageNet, each limited to a maximum of 100 iterations. The robust accuracy of these defense models under various attack scenarios was assessed to detail the outcomes.


\begin{table}[htbp]
\centering
\caption{The robustness (\%) of the models, evaluated using the \textbf{PMA} attack with varying $\beta$ values, on the CIFAR10 and ImageNet datasets.}
\begin{adjustbox}{width=1\linewidth}
\begin{tabular}{l|c|r|r|r|r|r}
\hline
\textbf{Dataset} & \textbf{Model} & $\boldsymbol{\beta = 0.5}$ & $\boldsymbol{\beta = 0.75}$ & $\boldsymbol{\beta = 1}$ & $\boldsymbol{\beta = 1.25}$ & $\boldsymbol{\beta = 1.5}$ \\
\hline
CIFAR10 & WRN-28-10\cite{cui2023decoupled} & 67.98 & 67.78 & \textbf{67.72} & 67.79 & 67.78 \\
CIFAR10 & WRN-28-10\cite{wang2023better} & 67.56 & 67.39 & \textbf{67.33} & 67.43 & 67.37 \\
CIFAR10 & RWRN-70-16\cite{peng2023robust} & 71.35 & 71.18 & \textbf{71.1} & 71.13 & 71.16 \\
ImageNet & ViT-B+CS\cite{singh2024revisiting} & 53.02 & 52.88 & \textbf{52.82} & 52.84 & 52.96 \\
ImageNet & Swin-B\cite{liu2023comprehensive} & 54.72 & 54.45 & \textbf{54.41} & 54.39 & 54.42 \\
ImageNet & ConvNeXt-S+CS\cite{singh2024revisiting} & 50.01 & 49.77 & \textbf{49.74} & 49.74 & 49.89 \\
\hline
\end{tabular}
\end{adjustbox}
\label{tab:pma_results}
\end{table}


For the parameter $\beta$, we investigated its influence across five distinct values: 0.5, 0.75, 1, 1.25, and 1.5. The results for the $PGD_{pm}$ method are detailed in Table \ref{tab:pgd_pm_results}, while those for the PMA method are presented in Table \ref{tab:pma_results}. The findings reveal distinct performance trends: the $PGD_{pm}$ method achieves marginally superior performance with $\beta = 0.75$, whereas the PMA method yields optimal results with $\beta = 1$.

\section{Million-Scale adversarial robustness evaluation between the PMA method and other methods}
 
In this supplementary section, we broaden our comparative analysis by incorporating the PMA and $PGD_{ce}$ methods with other existing techniques. We assessed the same set of five ImageNet defense models discussed in the main body of the paper. Due to the extensive duration—estimated to span several months—to test AA on the CC1M dataset, we chose not to conduct this test. Instead, to enhance our evaluation, we randomly selected a subset of 10,000 images from CC1M to evaluate the comparative robustness of AA. For consistency, we allocated 100 steps for all methods, with the exception of AA, which includes four distinct attacks.

The results of these experiments are presented in Table \ref{tab:model_robustness_results}, where we also document the computational time expended by one of the defense models when subjected to various attack methodologies. All experiments were conducted on an NVIDIA RTX 3090 GPU with a batch size of 32. As shown in Table \ref{tab:efficiency_results}, AA emerges as the superior approach; however, our PMA method closely matches AA in performance while requiring only 3\% of AA’s evaluation time.


\begin{table}[htpb]
\centering
\caption{The efficiency results (in seconds) across various methods on CC1M, with the best performances in bold.}
\begin{adjustbox}{width=1\linewidth}
\begin{tabular}{l|c|r|r|r|r|r|r|r|r|r}
\hline
\textbf{Dataset} & \textbf{Model} & \textbf{$\textbf{PGD}_\text{ce}$} & \textbf{$\textbf{PGD}_\text{dlr}$} & \textbf{$\textbf{PGD}_\text{mg}$} & \textbf{$\textbf{PGD}_\text{pm}$} & \textbf{$\textbf{PGD}_\text{alt}$} & \textbf{$\textbf{PGD}_\text{mi}$} & \textbf{$\text{MD}$} & \textbf{$\text{PMA}$} & \textbf{$\text{AA}$} \\
\hline
ImageNet & Swin-B\cite{liu2023comprehensive} & 766 & 700 & 742 & 742 & 688 & 878 & 742 & 718 & 22328 \\
\hline
\end{tabular}
\end{adjustbox}
\label{tab:efficiency_results}
\end{table}

\section{Supplementary experiments on CLIP}
To address the domain mismatch between Conceptual-Captions and ImageNet, we conducted supplementary experiments using CLIP on CC1M. We tested perturbation ranges of 1, 2, and 3, with a batch size of 32. All samples in a batch—except the text corresponding to the given image—were treated as negative samples. As shown in Table~\ref{tab:clip_attack}, our method consistently achieves the best performance.

\vspace{10pt}
\begin{table}[htbp]
    \centering
    \begin{adjustbox}{width=0.5\textwidth}
    \begin{tabular}{ccccccccccc}
        \toprule
        \textbf{$\epsilon$} & \textbf{Clean} & \textbf{PGD$_{ce}$} & \textbf{PGD$_{dlr}$} & \textbf{PGD$_{mg}$} & \textbf{PGD$_{pm}$} & \textbf{PGD$_{alt}$} & \textbf{PGD$_{mi}$} & \textbf{MD} & \textbf{PMA} &\textbf{diff}\\
        \midrule
        1/255 & 73.74 & 26.22 & 22.32 & 21.92 & 21.9 & 21.4 & 21.32 & 21.32 & \textbf{21.16} &\textbf{-0.16}\\
        2/255 & 73.74 & 13.12 & 10.45 & 9.97 & 9.94 & 9.48 & 9.5 & 9.5 & \textbf{9.4} &\textbf{-0.1}\\
        3/255 & 73.74 & 7.42 & 6.2 & 5.87 & 5.89 & 5.5 & 5.55 & 5.57 & \textbf{5.45} &\textbf{-0.05}\\
        \bottomrule
    \end{tabular}
    \end{adjustbox}
    \caption{Robustness (\%) evaluated under varying $\epsilon$.}
    \label{tab:clip_attack}
\end{table}

\end{document}